\documentclass{article}

\PassOptionsToPackage{numbers, compress}{natbib}

\usepackage[preprint]{neurips_2026}
\usepackage{titlesec} 
\usepackage{titletoc} 

\usepackage[utf8]{inputenc} 
\usepackage[T1]{fontenc}    
\usepackage{hyperref}       
\usepackage{url}            
\usepackage{booktabs}       
\usepackage{amsfonts}       
\usepackage{nicefrac}       
\usepackage{microtype}      
\usepackage{xcolor}           
\usepackage{graphicx}

\usepackage{multirow}   
\usepackage{enumitem}
\usepackage{tcolorbox}

\usepackage{amsmath}
\usepackage{amssymb}
\usepackage{mathtools}
\usepackage{amsthm}

\title{Beyond Average Performance: Dynamic Instance Clustering and Specialized Algorithm Design in LLM-Assisted Evolutionary Search}

\author{%
Qinglong Hu$^{1}$, Qingfu Zhang$^{1}$ ,Fei Liu$^{1}$, Xialiang Tong$^{2}$, Kun Mao$^{2}$ \& Mingxuan Yuan$^{2}$\\
$^{1}$Department of Computer Science, City University of Hong Kong, Hong Kong, China \\
$^{2}$Huawei Noah’s Ark Lab, China\\
\texttt{qinglhu2-c@my.cityu.edu.hk, qingfu.zhang@cityu.edu.hk} \\}

\begin{document}

\maketitle

\begin{abstract}
  Large Language Model-assisted Evolutionary Search (LES) has emerged as a powerful paradigm for automated algorithm design. However, existing LES methods primarily optimize for average performance, inherently directing search effort toward instances that contribute most to this metric while leaving others poorly served, resulting in weak tail robustness and limited real-world reliability. To address this limitation, we propose \textbf{\underline{Dy}}namic Instance \textbf{\underline{C}}lustering and Specialized \textbf{\underline{A}}lgorithm Design (DyCA), an LES framework with a feature-free, structure-aware mechanism for constructing reliable algorithm portfolios under heterogeneous instance distributions. DyCA treats instance clustering as a co-evolving component within the search process, reusing accumulated evaluation data as feature-free signals to progressively partition instances with similar algorithmic response patterns. The uncovered clusters decompose the mixed objective into a set of structure-aware sub-objectives, thereby enabling finer-grained and more adaptive guidance for specialized algorithm design. Experimental results across four algorithm design tasks with heterogeneous instances demonstrate that DyCA outperforms state-of-the-art LES baselines, improving tail robustness by an average of 15.2\% and overall performance by 7.1\% while maintaining competitive head performance. 
\end{abstract}

\section{Introduction}
Large Language Model-assisted Evolutionary Search (LES) has recently emerged as a powerful paradigm for automated algorithm design~\cite{liu2024systematic}. By integrating the generative reasoning capabilities of Large Language Models (LLMs) with the iterative optimization strengths of evolutionary computation, LES enables the automated synthesis of high-performing algorithms across diverse domains, including optimization~\cite{romera2024mathematical, ye2024reevo, dat2025hsevo, van2024llamea}, symbolic regression~\cite{shojaee2025llmsr}, and machine learning~\cite{mo2025autosgnn, zhou2025design, hu2025automated}.

Despite these successes, a fundamental limitation remains. Most existing LES frameworks optimize algorithms with respect to \emph{average performance} over a given set of problem instances~\cite{liu2024evolution}. However, in practice, real-world instance distributions are often highly heterogeneous. Under such conditions, optimizing for the mean objective inevitably induces a \emph{majority-dominance bias}: to maximize the average score, the search process tends to favor the majority or easier instances that contribute most to the aggregate score, while neglecting minority or more challenging ones~\cite{Sagawa2020Distributionally, lehman2011abandoning}. As a result, the algorithms designed often exhibit strong head or mean performance but poor tail performance. This lack of \textbf{tail robustness} limits the reliability of LES, particularly in mission-critical contexts where worst-case performance and robustness are essential~\cite{2005Jinyc}.

Several recent efforts have attempted to improve the reliability of LES under heterogeneous instance distributions. EoH-S~\cite{liu2025eohs} introduces a complementary algorithm pool to enhance coverage across instances. However, as its evolutionary guidance is driven by average gain, the evolution of the algorithm pool tends to prioritize instances that contribute most to the aggregate metric. From a more principled perspective, Instance Space Analysis (ISA)~\cite{Rasulo2024} provides a promising approach for addressing heterogeneity by partitioning instances into structurally distinct regions and enabling specialized algorithm design~\cite{zhang2025llm}. Unfortunately, classical ISA heavily relies on domain-specific instance features~\cite{munoz2018instance}. In the context of LES, which is frequently applied to novel or black-box domains, such features are often unavailable or poorly defined, rendering ISA impractical. This exposes a dilemma: mitigating the majority-dominance bias requires a structural understanding of the instance space, yet such structure is inaccessible without reliable prior features. This motivates our central research question: \textbf{How can LES be empowered to design algorithms with strong tail robustness over heterogeneous instances, in the absence of predefined instance features?}

\begin{figure}[ht]
    \includegraphics[width=1.0\textwidth]{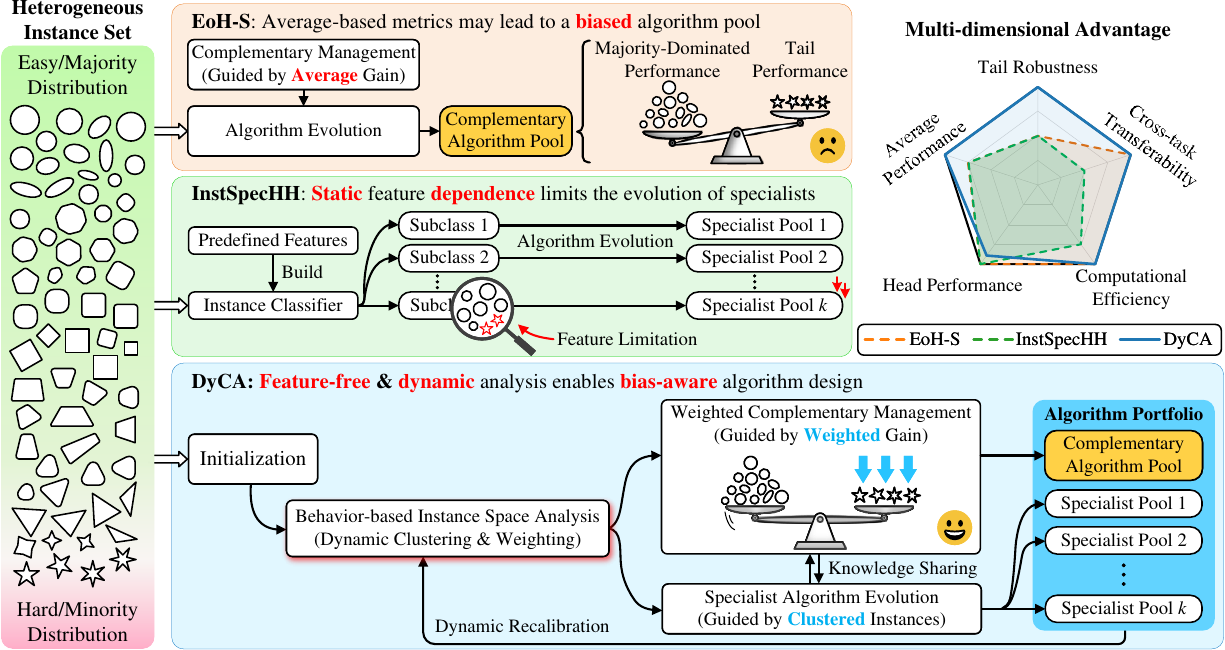}
    \caption{
      Conceptual comparison of LES paradigms under heterogeneous settings.
\textbf{Top:} EoH-S relies on average-guided complementary management, which induces majority-dominance bias.
\textbf{Middle:} InstSpecHH~\cite{zhang2025llm} alleviates bias via instance specialization but is constrained by predefined features.
\textbf{Bottom:} DyCA performs feature-free, behavior-based dynamic instance space analysis, enabling bias-aware specialized algorithm design.
The radar plot highlights DyCA’s multi-dimensional advantages, particularly in tail robustness.}
    \label{method_comparasion}
\end{figure}

To address this challenge, we propose \textbf{Dy}namic Instance \textbf{C}lustering and Specialized \textbf{A}lgorithm Design (\textbf{DyCA}), a feature-free LES framework designed to enhance reliability under heterogeneous instance distributions. Rather than relying on predefined instance features for offline ISA, DyCA adopts a behavior-based and dynamic perspective to characterize instance heterogeneity. It is grounded in a fundamental insight: the intrinsic difficulty and solvability of an instance can be naturally characterized by how it differentially responds to a diverse set of algorithms.

Building on this insight, DyCA repurposes the algorithm–instance evaluation data routinely accumulated during the LES process to perform progressive, online instance analysis. Instances exhibiting similar algorithmic response patterns are dynamically identified and clustered, uncovering the latent structure of the instance space. This behavior-induced structure enables DyCA to explicitly counteract majority-dominance bias by reallocating search efforts toward underserved or difficult instance groups, thereby facilitating the design of more effective specialized algorithms. By continuously interleaving instance-structure discovery with algorithm evolution, DyCA enables bias-aware resource allocation within LES and substantially improves tail robustness without compromising overall performance, thereby leading to more reliable algorithm portfolios. In summary, our contributions are as follows:

(1) We introduce \textbf{B}ehavior-based \textbf{D}ynamic \textbf{I}nstance {\textbf{S}pace \textbf{A}nalysis (B-DISA), a fully data-driven, low-overhead method that progressively uncovers latent instance structure from routine LES processes. B-DISA explicitly groups instances with similar solvability patterns, enabling fine-grained and efficient specialized algorithm design.

(2) We propose DyCA, a unified framework that tightly integrates B-DISA with LES through a coupled feedback loop. LES continuously supplies behavioral data for B-DISA, while B-DISA provides increasingly accurate structural awareness to guide specialized LES. This mutual reinforcement enables bias-aware algorithm evolution under heterogeneous instance distributions.

(3) We empirically evaluate DyCA on four algorithm design tasks. The results demonstrate that DyCA consistently achieves substantial improvements in tail robustness and overall performance over state-of-the-art LES baselines. Extensive ablation studies further validate the effectiveness of B-DISA and the specialized design mechanism.

\section{Behavior-based dynamic instance space analysis}
\label{main_bdisa}

\subsection{Motivation}

Recent advances in LES, such as InstSpecHH~\cite{zhang2025llm}, demonstrate that integrating a divide-and-conquer strategy can effectively mitigate majority-dominance bias via instance specialization. By using ISA to partition instances into distinct subsets and allocating dedicated search effort to each, such methods improve robustness compared to purely average-performance-driven optimization.

However, existing ISA-based methods suffer from two fundamental limitations in the context of LES. \textbf{First}, they heavily rely on predefined, domain-specific instance features. The quality of these features directly constrains the induced partition, imposing an implicit performance ceiling on subsequent specialization. In novel domains where priors are unavailable, this reliance makes classical ISA impractical. \textbf{Second}, ISA is typically performed as an offline preprocessing step before LES. Any suboptimal partition generated at this stage becomes irreversible, resulting in persistent misallocation of computational resources throughout the design process.

To overcome these challenges, we propose Behavior-based Dynamic Instance Space Analysis (B-DISA), \textbf{a feature-free, online analysis method tailored for LES}. B-DISA progressively uncovers the latent structure of the instance space during the LES process, supporting more adaptive and effective instance specialization.

\subsection{Algorithms as probes, response vectors as features} 

Rather than representing instances via static, handcrafted features, B-DISA characterizes them through their behavioral responses to algorithms. B-DISA repurposes the algorithm--instance evaluation data accumulated throughout the LES process. Let $\mathcal{I} = \{i_1, \dots, i_M\}$ denote the set of $M$ instances, and $\mathcal{A} = \{a_1, \dots, a_n\}$ the archive of algorithms generated up to the current evolutionary stage. Evaluating $\mathcal{A}$ on $\mathcal{I}$ produces an algorithm--instance performance matrix $\mathbf{X} \in \mathbb{R}^{M \times n}$, where $X_{m,j}$ denotes the performance of algorithm $a_j$ on instance $i_m$. The $m$-th row of $\mathbf{X}$ is the \emph{response vector} of instance $i_m$, capturing its behavioral responses across all algorithms.

The central insight of B-DISA is that an instance’s intrinsic difficulty and solvability are reflected in its response vector. Two instances are considered behaviorally similar if they exhibit similar performance responses across a diverse set of algorithms. In this view, algorithms serve as \emph{behavioral probes} of the instance space, while response vectors act as domain-agnostic representations. By clustering these response vectors, B-DISA identifies instance clusters that are susceptible to similar algorithmic strategies, thereby providing cohesive sub-objectives for specialized algorithm design.

However, directly clustering instances using full response vectors becomes increasingly impractical as the archive size $n$ grows during LES, resulting in high-dimensional representations. To address this, B-DISA introduces a compact representation using a small subset of \textbf{\emph{anchor algorithms}} $\mathcal{Z} \subset \mathcal{A}$, where $|\mathcal{Z}| \ll n$. These anchors are selected from the evolving algorithm archive, preserving instance discriminability and substantially reducing dimensionality. The resulting anchor-induced response matrix $\bar{\mathbf{X}} \in \mathbb{R}^{M \times |\mathcal{Z}|}$ provides a low-dimensional yet behaviorally informative embedding of the instance space for clustering. The detailed anchor selection mechanism is elaborated in Appendix~\ref{appendix_bdisa_anchor_selection}.

\subsection{Dynamic instance structure learning}

Unlike classical ISA, which assumes a fixed instance structure, B-DISA treats the instance structure as a latent variable that is progressively learned and refined during the LES process. As the anchor algorithm set becomes more diverse, newly observed response patterns uncover increasingly fine-grained behavioral distinctions between instances, further refining the inferred instance structure.

To support this dynamic refinement, B-DISA periodically recalibrates the instance structure based on updated behavioral evidence. Specifically, the anchor set is revised to incorporate algorithms that expose previously unobserved distinctions between instances, while redundant anchors are removed to maintain compactness (Appendix~\ref{appendix_recalibration}). Instances are then re-clustered using the updated anchor-induced response representations. This recalibration is lightweight and fully data-driven, reusing evaluation data already generated by LES and requiring no additional domain knowledge. As a result, early structural inaccuracies can be corrected rather than propagated throughout the search.

By continuously interleaving instance structure discovery with algorithm evolution, B-DISA provides LES with an adaptive, self-correcting view of the instance space, establishing a principled basis for bias-aware search allocation and specialized algorithm design.

\subsection{Properties and discussion}

B-DISA replaces static, feature-based, and offline ISA with a behavior-driven, dynamic, and online alternative, better suited for LES in heterogeneous and feature-scarce settings. \textbf{Computational efficiency:} It relies solely on evaluation data accumulated during LES and performs clustering on a compact, anchor-induced representation. \textbf{Generality:} It is entirely task-agnostic and domain-independent, supporting instance structure discovery for LES across a wide range of new tasks. It is also agnostic to the underlying LES methods and requires only evaluation data. \textbf{Robustness:} Although early-stage clusters may be coarse due to limited algorithm diversity, the dynamic recalibration mechanism ensures that structural accuracy improves as algorithm quality and diversity increase. 

\section{DyCA: Bias-aware LES framework integrating B-DISA}
\label{main_dyca}

By tightly incorporating B-DISA, we propose Dynamic Instance Clustering and Specialized Algorithm Design (DyCA), a reliability-oriented LES framework for automated algorithm design targeting heterogeneous instances. DyCA provides a unifying paradigm that enables LES methods to explicitly allocate search effort and regulate evolutionary pressure in a structure-aware manner, thereby improving the tail robustness of the resulting algorithm portfolio across heterogeneous settings. At a high level, DyCA operates as an iterative framework consisting of the following stages:

\textbf{1. Initialization:} DyCA begins by generating an initial algorithm pool through repeated prompting of LLMs. Previously generated algorithms are provided as contextual input in subsequent prompts to reduce redundancy. The primary purpose of this stage is to obtain an initial set of algorithm--instance performance responses, which serve as behavioral evidence to seed the B-DISA method.

\textbf{2. B-DISA assisted evolutionary cycle:} Each cycle consists of two tightly coupled steps:

\emph{(i) Dynamic instance structure learning:} Using the accumulated algorithm--instance evaluation data, B-DISA updates the set of anchor algorithms and recalibrates instance clusters based on anchor-induced behavioral response patterns. These clusters represent the current hypothesis of the latent instance structure and provide structural guidance for subsequent specialized algorithm design.

\emph{(ii) Bias-Aware Evolutionary Search:} Conditioned on the instance clusters yielded by B-DISA, DyCA modulates the evolutionary process in a structure-aware manner. Cluster information guides parent selection, population management, and the distribution of search effort and evolutionary pressure across the instance space. By assigning greater emphasis to poorly represented, hard-to-solve, or underserved clusters, DyCA systematically counteracts majority-dominance bias, which favors abundant or easy instances under conventional average-driven objectives. Consequently, the evolutionary search is steered toward producing algorithms that improve reliability across the entire instance space rather than optimizing performance on dominant subsets alone.

\textbf{3. Termination:} The cycle terminates when a predefined stopping criterion is met, such as reaching the maximum number of evaluated algorithms or exhausting the available LLM inference budget.

Importantly, DyCA is agnostic to the underlying evolutionary mechanisms, enabling easy integration with advanced LES methods. In this study, we instantiate DyCA by extending the core ideas of EoH-S and InstSpecHH within a unified evolutionary process, enabling the bias-aware co-evolution of both complementary and specialist algorithm pools. The framework is illustrated in Fig.~\ref{method_comparasion}.

\subsection{Weighted complementary population management}
Complementary Population Management (CPM) is the core mechanism of EoH-S for constructing a complementary algorithm pool that serves diverse instances~\cite{liu2025eohs}. Rather than selecting algorithms solely based on average performance across all instances, CPM promotes instance-wise complementarity by favoring algorithms that improve performance on instances that are not yet well served by the current algorithm pool. Concretely, given a pool $P_t$ and a candidate algorithm $\tilde{a}$, CPM evaluates its marginal contribution via the Complementary Performance Index (CPI):
\begin{equation}
\Delta \mathrm{CPI}(\tilde{a} \mid P_t) = \sum_{i_m \in \mathcal{I}} \max \left( f^*_{m}(P_t) - f_m(\tilde{a}), 0 \right),
\end{equation}
where $f^*_{m}(P_t)$ is the best-achieved score on instance $i_m$ by algorithms in $P_t$. By greedily selecting algorithms with the largest $\Delta \mathrm{CPI}$, CPM encourages pool-level complementarity across instances.

\textbf{Limitations in heterogeneous settings.} Although CPM promotes instance-wise complementarity, it implicitly assumes a uniform distribution of instances in the performance space. In heterogeneous scenarios, because $\Delta \mathrm{CPI}$ aggregates unweighted contributions over the entire instance set, the selection pressure is biased toward: (1) High-density clusters: Clusters containing a large number of behaviorally similar instances exert disproportionate influence on the aggregate gain; (2) High-gradient instances: Instances with larger absolute performance ranges (or ``easy-to-gain'' scores) dominate the selection process, even when they are already well served. 

Consequently, specialized algorithms tailored for minority clusters, which may represent critical edge cases or intrinsically difficult problem structures, are often overshadowed by algorithms that provide marginal improvements on the majority of instances. This prevents the evolution of a truly robust and complementary algorithm pool.

\textbf{Structure-aware weighted CPM (Ours).} To address this limitation, DyCA introduces a structure-aware variant termed Weighted CPM (W-CPM), which incorporates the instance clusters uncovered by B-DISA. Let $\{\mathcal{C}_1, \dots, \mathcal{C}_K\}$ denote the current partition of the instance set. Each instance $i_m \in \mathcal{C}_k$ is assigned a weight $1/|\mathcal{C}_k|$, yielding the weighted marginal contribution:
\begin{equation} 
\Delta \mathrm{CPI}_w(\tilde{a}{\mid}P_t){=}\sum^K_{k{=}1}\sum_{i_m{\in}\mathcal{C}_k}  \frac{1}{|\mathcal{C}_k|}{\cdot}\max\left(f^*_{m}(P_t){-}f_m(\tilde{a}),0 \right)
\end{equation} 
This weighting ensures that each instance cluster contributes equally to the selection objective, preventing large or easy clusters from overwhelming the evolutionary pressure. Notably, W-CPM generalizes the original CPM: when all instances belong to a single cluster, the two formulations coincide. Moreover, because the weighting preserves non-negativity and additivity, the greedy selection procedure retains the theoretical guarantees of CPM~\cite{liu2025eohs}, while substantially improving robustness under heterogeneous instance distributions. The proof is provided in Appendix~\ref{Theoretical_property_WCPM}.

\subsection{Bias-aware specialist algorithm design}
A fundamental challenge in algorithm design for heterogeneous instances lies in objective entanglement: when instances with incompatible solvability characteristics are optimized jointly, the resulting search signal becomes inherently conflicted. Such conflicts lead to inefficient use of the limited LLM inference budget, as substantial search effort is inevitably spent on contradictory exploration.

DyCA addresses this issue by leveraging the instance structure uncovered by B-DISA. By clustering instances based on the similarity of their algorithmic response patterns, the original heterogeneous optimization objective is decomposed into a set of more coherent sub-objectives. Instances within the same cluster exhibit similar solvability and tend to benefit from aligned algorithmic strategies, making each cluster a focused target for specialist algorithm design. Based on this structural decomposition, DyCA maintains a dedicated LES process for each cluster, with objectives restricted to improving performance on that cluster alone. Operating on these more coherent and targeted objectives improves the efficiency of specialized LES, enabling advanced LES techniques to be fully leveraged within each cluster-specific process to produce superior algorithm designs.

The explicitly uncovered instance structure further enables a bias-aware allocation of the limited LLM inference budget among instances. Rather than distributing search effort evenly, DyCA dynamically allocates resources across instance clusters based on their progress toward specific sub-objectives. Clusters that are already progressing well can be deprioritized, while those that are underperforming or are inherently difficult to solve receive more attention. This bias-aware approach ensures that the specialized evolution consistently targets the most underserved areas of the instance space. More details of the bias-aware allocation strategy are provided in Appendix~\ref{appendix_specialist_pool}.

\subsection{Resource allocation and knowledge sharing in DyCA}

DyCA organically integrates complementary and specialist evolution through coordinated resource allocation and cross-pool knowledge sharing, ensuring that global robustness and local specialization are jointly optimized throughout the search process.

\textbf{Adaptive resource allocation.} In early evolutionary stages, when the instance structure inferred by B-DISA is still coarse or unstable, DyCA prioritizes the complementary pool to promote broad instance coverage and accumulate informative behavioral evidence. As instance clusters stabilize across successive recalibration steps, an increasing proportion of the search effort is shifted toward specialist pools, enabling sustained and focused optimization within each cluster.

\textbf{Knowledge sharing.} All newly generated algorithms are evaluated on the full instance set and made accessible to both complementary and specialist pools. This shared evaluation and archive protocol facilitates rapid dissemination of valuable algorithmic innovations across diverse pools. In addition, when a specialist pool exhibits prolonged stagnation, DyCA allows it to selectively incorporate algorithms from the complementary pool or from other specialist pools, introducing structurally diverse search signals to facilitate escape from local optima.

\section{Experiments}
\label{sec:experiments}

\subsection{Experimental Settings}
\label{sec:exp_settings}

\textbf{Algorithm design tasks} We evaluate DyCA on four automated algorithm design tasks. 
Following recent advances in LES~\cite{ye2024reevo, liu2025eohs, zhang2025llm}, we consider the Traveling Salesman Problem (TSP), Capacitated Vehicle Routing Problem (CVRP), and Online Bin Packing (OBP), enabling a direct comparison. Additionally, we include a Lunar Lander Control (LLC) task, which involves designing control policies~\cite{hu2025mles}, to assess the transferability of DyCA beyond common algorithm design scenarios.

For TSP and CVRP, DyCA designs constructive and routing heuristics, respectively, and performance is measured by the optimality gap relative to the LKH-3 solver~\cite{helsgaun2017extension}.
For OBP, DyCA designs online packing strategies and is evaluated by the excess ratio over a theoretical lower bound~\cite{martello1990lower}.
For LLC, DyCA designs control policies that map state vectors to thrust actions, and performance is measured by cumulative reward, with scores above $200$ indicating successful task completion~\cite{hu2025mles}.

\textbf{Heterogeneous instance sets.} To rigorously assess reliability under instance heterogeneity, we construct highly diverse training and testing instance sets for each task. 
For TSP, CVRP, and OBP, each training set consists of 240 instances spanning various problem scales and multiple generative distributions, with at least three distributions per task. This setup introduces variability in instance difficulty and requires different algorithmic strategies to achieve effective solutions.
The LLC training set contains 35 instances covering a wide range of initial states with varying difficulty. \textbf{Testing instances} are sampled from the same underlying distributions as the training sets. The testing sets contain 120 instances for TSP, CVRP, and OBP, and 50 instances for LLC. Detailed task descriptions, instance compositions, and distribution statistics are provided in Appendix~\ref{appendix_tasks}.

\textbf{Baselines.} We compare DyCA against two categories of state-of-the-art methods:
(i) standard LES methods, including EoH~\cite{liu2024evolution}, FunSearch~\cite{romera2024mathematical}, and ReEvo~\cite{ye2024reevo}, which aim to design a single algorithm optimized for average performance;
and (ii) reliability-enhanced LES methods, including InstSpecHH~\cite{zhang2025llm} and EoH-S~\cite{liu2025eohs}, which explicitly account for instance diversity.

\textbf{Implementation details.}
All methods employ the GPT-4o-mini LLM with default generation parameters to generate offspring algorithms. For fairness, each method is allocated an identical budget of 2{,}000 LLM inference queries on each task. The algorithm pool size is set to 10 for single-pool methods (EoH, ReEvo, EoH-S). For multi-pool methods (DyCA and InstSpecHH), each sub-pool also contains 10 individuals, ensuring comparable local search intensity. InstSpecHH relies on predefined features to cluster the instances. To simulate an idealized feature setting and avoid confounding effects, we provide InstSpecHH with ground-truth cluster labels. In contrast, all other methods operate in a strictly feature-free setting. All reported results are averaged over three independent runs using the same training instance sets to mitigate stochastic effects.

\textbf{Evaluation protocol.}
All methods are evaluated following their recommended protocols. EoH, FunSearch, and ReEvo evaluate a single algorithm selected based on the best average performance on the training set. EoH-S evaluates its complementary algorithm pool as the final output. For a controlled comparison, DyCA adopts the same protocol and reports the performance of its complementary pool. For InstSpecHH, we evaluate the union of the best-performing algorithms across all specialist pools, thereby avoiding potential performance degradation from instance-to-specialist matching errors~\cite{zhang2025llm}.

\subsection{Comparative evaluations} 

Table~\ref{tab:average_tail} summarizes the comparative performance of all methods on the testing instances across four tasks, reporting both the average performance and the performance on the worst 10\% of instances (\textit{Tail}). The tail instances are identified separately for each method based on its own performance ranking. Arrows ($\downarrow$ / $\uparrow$) indicate whether lower or higher values are preferred. The best results are highlighted in \textbf{bold}, and the second-best results are \underline{underlined}. Figure~\ref{fig:box_2of4} provides an additional intuitive analysis of performance distributions across instances for the CVRP and TSP tasks.

\begin{table}[h]
\centering
\caption{Comparison of average and tail performance on testing instances.}
\label{tab:average_tail}
\resizebox{\linewidth}{!}{
\begin{tabular}{lcccccccc}
\toprule
\multirow{2}{*}{Method} & \multicolumn{4}{c}{Average Performance}                                                     & \multicolumn{4}{c}{Tail Performance}                                                        \\ \cmidrule(lr){2-5} \cmidrule(lr){6-9}
                        & CVRP($\downarrow$) & TSP ($\downarrow$) & OBP ($\downarrow$) & LLC ($\uparrow$) & CVRP($\downarrow$) & TSP ($\downarrow$) & OBP ($\downarrow$) & LLC ($\uparrow$) \\ \midrule
EoH                     & 0.2106             & 0.1346             & 0.0278             & 200.68           & 0.4561             & 0.2324             & 0.0553             & 91.92            \\
Funsearch               & 0.2273             & 0.1373             & 0.0277             & 216.14           & 0.4884             & 0.2503             & 0.0548             & 88.19            \\
ReEvo                   & 0.2241             & 0.1331             & 0.0281             & 242.09           & 0.4548             & 0.2348             & 0.0555             & 158.71           \\
EoH-S                   & \underline{0.1489}       & 0.0815       & \underline{0.0233}       & \underline{270.55}     & \underline{0.3332}       & 0.1588      & \underline{0.0431}       & 167.01           \\
InstSpecHH              & 0.1617             & \underline{0.0778}       & 0.0236       & 268.22     & 0.3651             & \underline{0.1552}       & 0.0469             & \underline{179.96}     \\
DyCA                    & \textbf{0.1361}    & \textbf{0.0738}    & \textbf{0.0227}    & \textbf{285.02}  & \textbf{0.2890}    & \textbf{0.1406}    & \textbf{0.0421}    & \textbf{220.55}  \\ \bottomrule
\end{tabular}
}
\end{table}

\vskip -0.1in
\begin{figure}[h]
  \begin{center}
    \centerline{\includegraphics[width=0.96\textwidth]{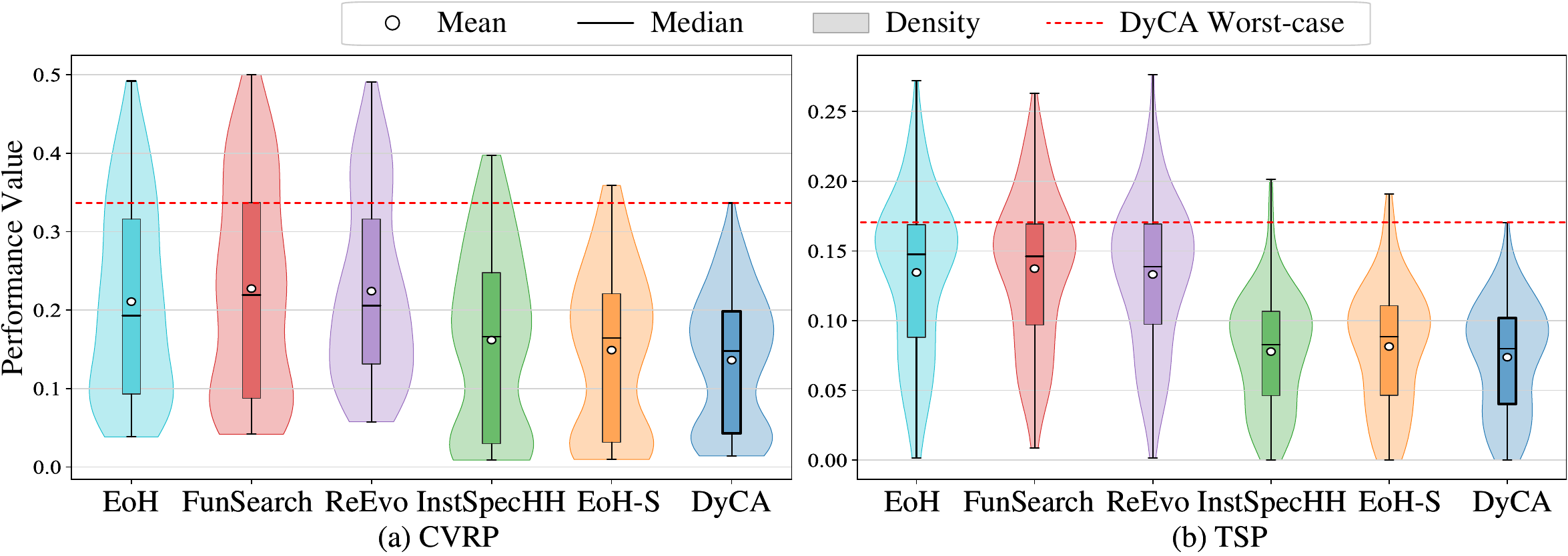}}
    \caption{
    Performance distributions across testing instances visualized using violin plots with embedded box plots. Panels (a) and (b) correspond to CVRP and TSP. 
    The shaded regions represent density estimates, and the dashed red line indicates the worst-case performance achieved by DyCA. 
    }
    \label{fig:box_2of4}
  \end{center}
   \vskip -0.2in
\end{figure}

\textbf{Overall performance.} As shown in Table~\ref{tab:average_tail}, DyCA consistently achieves the best average performance across all four tasks. Averaged over all tasks, DyCA improves performance by \textbf{33.2\%} compared to standard LES methods and by \textbf{7.1\%} over state-of-the-art reliability-enhanced baselines. These results demonstrate the general superiority of DyCA across diverse algorithm design tasks.

\textbf{Tail robustness.} 
A central objective of DyCA is to enhance tail robustness, thereby improving reliability under heterogeneous conditions. The right half of Table~\ref{tab:average_tail} reports performance on the worst 10\% of testing instances. DyCA exhibits a clear and consistent advantage in this regime. Compared to InstSpecHH and EoH-S, DyCA improves tail performance by an average of \textbf{15.2\%} across the four tasks, and by \textbf{53.1\%} relative to standard LES methods. Notably, on the LLC task, DyCA is the only method that achieves a tail score exceeding 200, indicating successful task completion even under the most challenging conditions. This result highlights DyCA’s ability to provide strong stability and reliability guarantees that are critical for real-world deployment.

\textbf{Head performance.} Importantly, the improvement in tail robustness does not come at the expense of head performance. As illustrated in Figure~\ref{fig:box_2of4}, DyCA maintains highly competitive performance on the best-performing instances while improving worst-case behavior. More comprehensive results and detailed quantitative analysis of the performance distribution are provided in Appendix~\ref{appendix_tail_analysis}.

Overall, DyCA significantly enhances tail robustness without sacrificing head performance, resulting in a more balanced and reliable algorithm portfolio across heterogeneous instances.

\subsection{Generalization results on public benchmarks}

We further assess the generalization of DyCA on public benchmarks, including CVRPLib~\cite{uchoa2017new}, TSPLib~\cite{reinelt1991tsplib}, and BPPLib~\cite{delorme2018bpplib}. These benchmarks consist of real-world or widely adopted instances with diverse characteristics, providing a complementary evaluation of practical effectiveness.

\begin{table}[h]
\caption{Performance comparison on standard public benchmarks.}
\label{tab:benchmark}
\begin{tabular*}{\linewidth}{c@{\extracolsep{\fill}}cccc}
\toprule
\textbf{Methods} & \textbf{CVRPLib} & \textbf{TSPLib} & \textbf{BPPLib} \\ \midrule
InstSpecHH & 0.3187          & 0.1088          & 0.1109          \\
EoH-S      & 0.2772          & 0.1127         & 0.1100         \\
DyCA       & \textbf{0.2448} & \textbf{0.1060} & \textbf{0.1084} \\ \bottomrule 
\end{tabular*}
\end{table}

As shown in Table~\ref{tab:benchmark}, DyCA consistently achieves the best performance across all benchmarks. Importantly, these instances are unseen during LES and may follow distributions substantially different from the constructed instance sets. Despite this distributional shift, DyCA consistently outperforms state-of-the-art LES methods, indicating improved generalization in practical settings. 

\subsection{Effectiveness of B-DISA.} 
A key hypothesis of DyCA is that B-DISA can identify and organize heterogeneous instances solely from algorithmic response patterns, without handcrafted features or prior distributional knowledge, and that the resulting structure can effectively guide bias-aware specialized algorithm design.

\begin{figure*}[h]
  \begin{center}
    \centerline{\includegraphics[width=\textwidth]{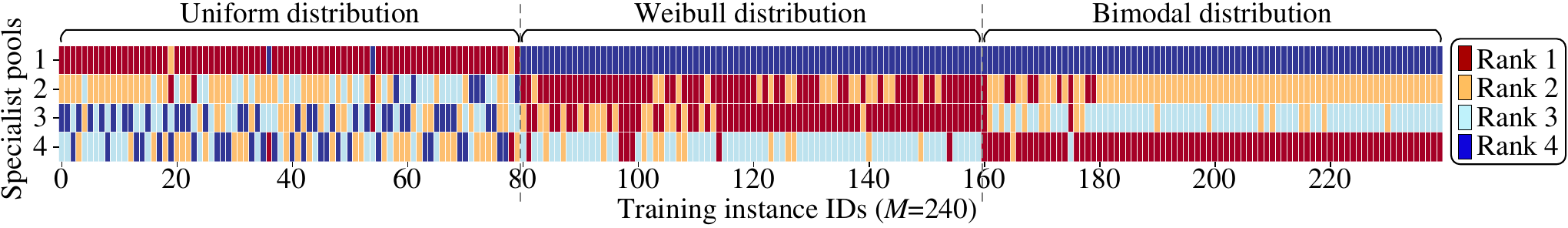}}
    \caption{
    Performance ranks of specialist pools on OBP training instances. The x-axis shows the 240 training instances, grouped by their ground-truth distribution types (Uniform, Weibull, Bimodal), while the y-axis corresponds to the four evolved specialist pools. 
    }
    \label{fig:heatmap}
  \end{center}
  \vskip -0.1in
\end{figure*}

To empirically validate this, we visualize the performance rankings of the evolved specialist pools on the OBP training set in Fig.~\ref{fig:heatmap}. The heatmap exhibits a clear block-wise structure, indicating that different pools have specialized to dominate distinct regions of the instance space. Specifically, Pool~1 achieves the best rank predominantly on instances drawn from the \textit{Uniform} distribution, Pools~2 and~3 show strong dominance on \textit{Weibull}, while Pool~4 consistently dominates the \textit{Bimodal}. 
Crucially, DyCA remains agnostic to these ground-truth labels throughout training. The observed specialization emerges purely from behavioral signals induced by algorithm-instance evaluation data. This provides strong empirical evidence that B-DISA organizes heterogeneous instances into behaviorally coherent clusters, enabling the spontaneous emergence of specialized algorithms without supervision.

Beyond its effectiveness, B-DISA is designed to be computationally lightweight. It operates by strictly repurposing the data accumulated during the routine LES loop. The operations for updating anchor algorithms and recalibrating instance clusters are computationally inexpensive, requiring less than two seconds empirically. This cost is infinitesimal compared to the dominant bottlenecks of LES: LLM-driven offspring generation (approx. 10--20s per query) and algorithm evaluation on instances.

In conclusion, B-DISA is a high-return, low-cost method that enhances DyCA's robustness and specialization capabilities, while corroborating the insight that purely behavior-based signals are sufficient to encode instance heterogeneity. More empirical analyses further examining the semantic meaning and convergence dynamics of the induced instance clusters are provided in Appendix~\ref{appendix:bdisa_insights}.

\subsection{Ablation study}
To quantify the contribution of individual components in DyCA, we conduct an ablation study on CVRP. The main results are summarized in Table~\ref{tab:ablation}, with additional results and analyses provided in Appendix~\ref{Additional_experimental_results}.

\begin{table}[h]
\caption{Ablation studies on CVRP.}
\label{tab:ablation}
\begin{tabular*}{\linewidth}{l@{\extracolsep{\fill}}ccc}
\toprule
\textbf{No.} & \textbf{Setting}        & \textbf{Mean}    & \textbf{Degradation} \\ \midrule
1            & DyCA (Full)                   & \textbf{0.1361}    & -    \\ \midrule
2            & No Clustering           & 0.1530          & 12.46 \%            \\
3            & Random Clustering       & 0.1495          & 9.89 \%              \\
4            & Feature Clustering (Oracle)      & 0.1366          & 0.35 \%              \\ \midrule
5            & w/o Weighting Mechanism & 0.1436 & 5.55 \%              \\
6            & w/o Specialist Evolution          & 0.1470          & 8.06 \%              \\ \bottomrule        
\end{tabular*}
\end{table}

\textbf{Importance of B-DISA.} Variants~2--4 investigate the role of ISA in enabling effective specialization.
\begin{itemize}[leftmargin=*, itemsep=3pt, topsep=0pt, parsep=0pt]
\item \textbf{No.~2 (No Clustering)} removes ISA entirely, reducing DyCA to an average-guided LES process. The substantial performance degradation ($12.46\%$) highlights the limitation of average-based guidance when confronted with highly heterogeneous instance distributions.
\item \textbf{No.~3 (Random Clustering)} randomly partitions instances into four clusters, simulating naive parallel evolution without meaningful instance organization. Although this variant improves upon No.~2, it still suffers a significant degradation ($9.89\%$), indicating that effective specialization requires structured and semantically meaningful clustering rather than arbitrary partitioning.
\item \textbf{No.~4 (Feature Clustering, Oracle)} clusters instances using ground-truth distribution labels available during instance generation, representing an oracle setting with ideal prior knowledge. DyCA achieves comparable and slightly better performance, suggesting that B-DISA can induce instance clustering that is competitive with expert-crafted, feature-based ISA for guiding specialization.
\end{itemize}

\textbf{Effectiveness of the weighting mechanism.} Variant~5 disables the weighted-CPM by assigning uniform weights to all instances, resulting in a 5.55\% performance drop. This demonstrates that structure-aware weighting is crucial for preserving and promoting algorithms tailored to difficult or underrepresented instances, thereby improving the overall robustness of the algorithm portfolio.

\textbf{Efficacy of bias-Aware specialist design.} 
Variant~6 removes the specialist pools and relies solely on the complementary pool for algorithm design. The resulting 8.06\% performance degradation demonstrates the importance of maintaining dedicated specialist pools. By explicitly evolving specialists on algorithmically compatible subsets of instances, DyCA enables more focused and efficient design than a single, globally guided pool.

\section{Conclusion}

This work focuses on a critical reliability gap in automated algorithm design using LLM-assisted Evolutionary Search (LES). We introduce Dynamic instance Clustering and specialized Algorithm design (DyCA), a novel framework designed to effectively handle heterogeneous instance distributions. DyCA leverages naturally accumulated data during LES to progressively uncover latent instance structures in a completely feature-free manner, enabling bias-aware evolution of specialist algorithms across the instance space. Extensive experiments demonstrate that DyCA significantly enhances tail robustness while consistently improving overall performance, highlighting its promise as a general and principled approach for reliable automated algorithm design in complex, real-world settings.



\medskip
{\small
\bibliographystyle{ieeetr}
\bibliography{nips_dyca}              
}
\newpage
\appendix

\section*{\centering \LARGE Appendix}

\textbf{\Large Contents}

\addcontentsline{toc}{section}{Appendix} 

\vspace*{5pt} 

\titlecontents{section}[2.5em] 
{\addvspace{5pt}\bfseries}      
{\contentslabel{2em}}           
{\hspace*{-2em}}                
{\titlerule*[0.7pc]{.}\contentspage} 

\titlecontents{subsection}[5.0em] 
{\addvspace{2pt}}                
{\contentslabel{2.5em}}         
{\hspace*{-2.5em}}              
{\titlerule*[0.7pc]{.}\contentspage} 

\startcontents[sections]

\printcontents[sections]{l}{1}{\setcounter{tocdepth}{3}}

\section{Related work}

\subsection{LLM-assisted evolutionary search}

LES employs LLMs as generative operators within evolutionary processes to iteratively design and refine algorithms based on empirical performance feedback~\cite{ael}. This paradigm has proven effective in various domains~\cite{ccetinkaya2025discovering, zheng2025cst, ye2025large, li2025llm}. Recent advances have primarily focused on enhancing search efficiency by integrating refined prompt engineering and mature evolutionary computation techniques into LES~\cite{ye2024reevo, xie2025llm, zheng2025monte, hu2025partition}. However, these methods are developed and evaluated based on average performance across training instances, paying limited attention to instance heterogeneity. Consequently, the search process is prone to a majority-dominance bias, which can undermine reliability. In contrast, our work targets reliability-oriented LES by explicitly modeling instance heterogeneity. By incorporating B-DISA into the evolutionary loop, DyCA enables bias-aware allocation of search effort, thereby improving tail robustness in heterogeneous settings.

\subsection{Instance space analysis}

ISA characterizes the relationship between problem structure and algorithm performance by embedding instances into a measurable space~\citep{Smith2023}. By revealing latent instance structure, ISA has been successfully applied to instance generation~\cite{SMITHMILES2015102, yap2022informing}, algorithm selection~\cite{rice1976algorithm, strassl2022instance}, and, more recently, algorithm design~\citep{liu2019automatic, zhang2025llm}. However, conventional ISA relies on domain-specific instance features, limiting its applicability with LES, which often targets novel domains where features are poorly defined. To address this, we propose B-DISA, which reformulates ISA as a feature-free and dynamic process. Instead of static priors, B-DISA infers instance similarity from accumulated algorithmic response patterns observed during the LES process. This enables instance structures to be refined online and fed back into LES, supporting more accurate specialized algorithm design.


\section{Limitations, discussion, and future work}
\label{appendix_discussion}

\paragraph{Limitations of response-based representation.}
DyCA uses performance responses as domain-agnostic features to analyze latent instance structures. This design imposes certain requirements on the response metric. Specifically, the response of an algorithm on the same instance across multiple runs must be both stable and accurate. Otherwise, noisy or inconsistent responses may propagate through the B-DISA process, distort the identification of instance structures, and consequently undermine the effectiveness of the subsequent specialized algorithm design in DyCA.

A representative counterexample is the use of algorithm's runtime as a response metric. In practice, runtime is often sensitive to external factors such as hardware parallelism and resource contention. The same algorithm on the same instance may exhibit significantly different runtime under low-load versus high-load conditions, leading to substantial inconsistency. 

Therefore, when applying DyCA, it is crucial to ensure that the response metric exhibits \emph{consistency and robustness}, and to prefer metrics that are less sensitive to environmental variations (e.g., solution quality, optimality gap, normalized objective values, or constraint violation measures). These types of metrics are widely available across combinatorial optimization, planning, and scheduling problems, indicating that the applicability of DyCA remains broad despite this requirement.

\paragraph{Inference cost and reliability trade-off.} DyCA is designed to yield robust and reliable algorithm portfolios by explicitly counteracting instance heterogeneity and majority-dominance bias in the LES process. This reliability is achieved through the specialization and coordinated deployment of multiple algorithms, rather than reliance on a single universally optimal solution. This design choice is well aligned with the No-Free-Lunch principle~\cite{wolpert1995no, wolpert2002no}, which suggests that no single algorithm can perform optimally across all instance distributions.

However, this design also introduces an inherent trade-off. A limitation of DyCA, shared with other portfolio-based LES methods such as EoH-S, lies in its inference procedure. At test time, a new instance must interact with the learned portfolio to determine the most suitable algorithm, which incurs additional inference cost compared to single-algorithm approaches.

Importantly, this cost represents a deliberate and transparent trade-off rather than a drawback. DyCA is particularly well-suited for scenarios where robustness and reliability are prioritized over minimal inference latency, such as offline optimization, scheduling, and planning problems in which solutions are executed repeatedly or in parallel. In these settings, the inference overhead is amortized or negligible relative to execution cost, while the benefits in tail performance and stability are substantial.

\paragraph{Future work: toward ultra-fast inference.}
A promising direction for future work is to substantially accelerate inference via lightweight instance-to-pool matching mechanisms. Notably, the DyCA portfolio naturally supports multiple deployment granularities, ranging from coarse-grained complementary pools to fine-grained specialist pools. With an appropriate matching mechanism, it may be possible to solve a new instance with a single targeted algorithm evaluation.

One potential avenue is to transition \emph{from implicit behavioral clustering to explicit feature matching}. As discussed in Appendix~\ref{appendix:bdisa_insights}, although DyCA discovers instance clusters implicitly through behavioral responses, these clusters can serve as high-quality supervisory signals for learning explicit decision rules. Concretely, the clusters induced by B-DISA can be analyzed post hoc to identify interpretable structural characteristics (e.g., graph motifs in TSP, bin tightness in OBP, or spatial dispersion patterns in routing problems) that correlate strongly with each specialist pool. This would enable the construction of an \emph{Instance--Pool Matcher}: a lightweight classifier or heuristic rule that maps explicit instance features directly to the most suitable specialist pool. Such a mechanism could enable near-instant inference while retaining the specialization benefits of the DyCA portfolio.

Beyond heuristic matching, future research may explore hybrid strategies that combine coarse feature-based routing with selective behavioral probing, adaptive portfolio pruning under resource constraints, or continual refinement of instance--pool associations as new instances are encountered. Together, these directions point toward extending DyCA to real-time scenarios without compromising its robustness advantages.

\section{Understanding Majority-Dominance Bias in LES}
\label{appendix:majority_bias}

In this section, we provide concrete and intuitive examples to illustrate how majority-dominance bias arises in practice and why addressing it is essential for designing robust algorithm portfolios.

We identify two primary drivers of majority-dominance bias: 
(i) distributional imbalance in the instance space, and 
(ii) heterogeneity in the instance-level performance landscape. 
Although conceptually distinct, these two effects often co-occur and mutually reinforce each other in heterogeneous settings.

\textbf{Example 1: Dominance of high-density instance clusters.}
In many problem domains, the instance space is unevenly distributed, with a large fraction of instances exhibiting similar behavioral patterns. During the algorithm design process, improvements on such high-density clusters yield consistent and substantial gains to the average objective, thereby exerting disproportionate selection pressure. Consequently, algorithms that perform well on these majority clusters are repeatedly reinforced, even if they offer little or no improvement on minority clusters. Conversely, algorithms specialized for small but structurally distinct instance groups contribute marginally to the aggregate score and are therefore unlikely to survive selection. This mechanism suppresses the emergence of specialists targeting rare yet critical instance types.

\textbf{Example 2: Dominance of high-gradient (``easy-to-gain'') instances.}
Selection bias also arises from disparities in performance sensitivity across instances. Some instances exhibit large absolute performance ranges, making them ``easy to improve'' even when they are already well-served. Incremental gains on such instances can dominate the aggregate objective, overshadowing harder instances whose potential improvements are smaller in magnitude but more meaningful for overall robustness. In extreme cases, the attainable performance gains on easy instances can exceed the maximum achievable score on difficult ones, rendering the latter effectively invisible under average-based selection objectives. As a result, the search process continues to optimize already-solved instances, while difficult or edge-case instances receive insufficient evolutionary pressure.

Collectively, these two effects bias the evolutionary dynamics toward algorithms that maximize average performance, rather than those that provide complementary coverage across heterogeneous instance groups.

\paragraph{Empirical illustration.}
Figure~\ref{fig:bias_lead_to_what} presents an instance-level comparison of DyCA against two state-of-the-art LES baselines, EoH-S and InstSpecHH, on 120 CVRP testing instances. Each radial spoke corresponds to a single instance, with performance values plotted relative to the target objective value. The outer color-coded ring indicates the dominant algorithm on each instance.

\begin{figure}[h!]
  \begin{center}
    \centerline{\includegraphics[width=0.8\columnwidth]{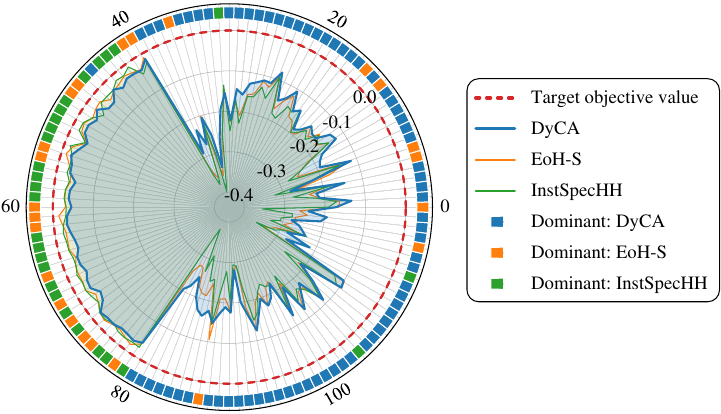}}
    \caption{Instance-level performance comparison of DyCA against EoH-S and InstSpecHH on 120 CVRP testing instances.
    For clearer visualization, performance scores are negated (values closer to the outer edge indicate better performance).
    Each radial spoke corresponds to a single test instance.
    The outer color-coded ring identifies the dominant algorithm for each instance (Blue: DyCA; Orange: EoH-S; Green: InstSpecHH),
    highlighting the tail-performance degradation induced by majority-dominance bias in baseline methods.
    }
    \label{fig:bias_lead_to_what}
  \end{center}
  \vskip -0.2in
\end{figure}

As shown in Figure~\ref{fig:bias_lead_to_what}, EoH-S and InstSpecHH exhibit intense competition and strong head performance on instances where all methods achieve high scores (e.g., Instances 40--80). However, their performance degrades substantially on more challenging instances (e.g., Instances 30--40 and 80--90), leading to pronounced tail performance failures. These patterns provide direct empirical evidence of majority-dominance bias and its detrimental impact on robustness.

We further analyze the underlying causes of this behavior. EoH-S constructs its complementary pool based on average performance gain, making it inherently vulnerable to high-gradient bias. Although InstSpecHH avoids average-based selection, it distributes the limited number of LLM queries uniformly across the instance space. Since difficult cases typically require substantially more search effort to solve, this uniform allocation results in insufficient attention to hard cases while allocating excessive resources to already-easy ones.

In contrast, DyCA exhibits a markedly more uniform and robust profile across the entire instance set. It maintains competitive parity in high-performing regions while demonstrating superior resilience in low-performing ones. By explicitly discovering instance heterogeneity via B-DISA, DyCA assigns implicit importance weights to instances during complementary pool construction and performs bias-aware resource allocation within specialist pools. Consequently, the evolutionary process actively counteracts majority-dominance bias, thereby enabling the emergence of a truly complementary and robust algorithm portfolio in heterogeneous settings.

\section{Extended experimental results}
\label{Additional_experimental_results}

\subsection{Detailed experimental setup}
\label{detailed_exp_setting}

This subsection reports the detailed experimental configuration and hyperparameter settings used in all experiments. All methods were implemented in Python and executed on a single CPU (Intel i7-13700) with 64 GB of RAM. Table~\ref{parametersetting_LES} summarizes the shared hyperparameters for DyCA and all baseline methods. Unless otherwise specified, identical settings were used across methods to control for confounding factors and ensure a fair comparison.

\begin{table}[h]
\centering
\caption{Shared hyperparameter settings for all LES methods.}
\begin{tabular}{ll}
\toprule
\textbf{Parameter Description}                                  & \textbf{Value}  \\ \midrule
Pool size                                                        & 10              \\
LLM version used in the evolutionary operators              & GPT-4o-mini     \\
LLM sampling temperature  & 1               \\
Maximum LLM queries & 2,000 \\
B-DISA recalibration interval & Every 100 LLM queries           \\ \bottomrule
\end{tabular}
\label{parametersetting_LES}
\end{table}

For each task, DyCA is first used to initialize the algorithm pool to a size of 10. This initialized pool is then reused as the starting seed for every run of all LES methods. By fixing the initial pool across methods and runs, we explicitly eliminate the influence of initialization quality on the observed performance differences, allowing the comparison to focus solely on the effectiveness of the evolutionary and search strategies.

Due to the high computational cost of LLM-assisted evolutionary search, we follow prior work~\cite {liu2025eohs, yao2025multi} and report results from three independent runs to ensure robustness and reliability.

\subsection{Empirical stability and instance-level statistical significance}
\label{appendix_stats}

To evaluate the stability and reproducibility of the reported results, we further analyze performance variability across independent runs. Table~\ref{tab:ci95} reports the mean performance together with $95\%$ confidence intervals, computed as $Mean \pm 1.96 \cdot SEM$, where $SEM$ denotes the standard error of the mean across runs.

\begin{table}[h]
\centering
\caption{Performance comparison across three independent runs.
Results are reported as $Mean \pm 1.96 \cdot SEM$ (95\% confidence interval).
Lower values indicate better performance for CVRP, TSP, and OBP, while higher values are better for LLC.
Best results are shown in bold.}
\label{tab:ci95}
\begin{tabular}{lcccc}
\toprule
\multicolumn{1}{l}{\textbf{Method}} & \multicolumn{1}{c}{\textbf{CVRP ($\downarrow$)}} & \multicolumn{1}{c}{\textbf{TSP ($\downarrow$)}} & \multicolumn{1}{c}{\textbf{OBP ($\downarrow$)}} & \textbf{LLC ($\uparrow$)} \\ \midrule
\textbf{EoH}                        & 0.2106 $\pm$ 0.0128                                  & 0.1346 $\pm$ 0.0069                                 & 0.0278 $\pm$ 0.0003                                 & 200.68 $\pm$ 6.93                 \\
\textbf{Funsearch}                  & 0.2273 $\pm$ 0.0067                                  & 0.1373 $\pm$ 0.0029                                 & 0.0277 $\pm$ 0.0002                                 & 216.14 $\pm$ 28.02                \\
\textbf{ReEvo}                      & 0.2241 $\pm$ 0.0053                                  & 0.1331 $\pm$ 0.0059                                & 0.0281 $\pm$ 0.0001                                 & 242.09 $\pm$ 14.19               \\
\textbf{EoH-S}                      & 0.1489 $\pm$ 0.0052                                  & 0.0815 $\pm$ 0.0015                                 & 0.0233 $\pm$ 0.0002                                 & 270.55 $\pm$ 2.00                 \\
\textbf{InstSpecHH}                 & 0.1617 $\pm$ 0.0052                                 & 0.0778 $\pm$ 0.0070                                & 0.0236 $\pm$ 0.0008                                 & 268.22 $\pm$ 16.77                \\
\textbf{DyCA}                       & \textbf{0.1361 $\pm$ 0.0077}                         & \textbf{0.0738 $\pm$ 0.0035}                        & \textbf{0.0227 $\pm$ 0.0005}                        & \textbf{285.02 $\pm$ 7.19}        \\ \bottomrule  
\end{tabular}
\end{table}

Across all four tasks, DyCA consistently achieves the best average performance. Moreover, the variability across independent runs remains low, indicating stable optimization behavior despite the inherent stochasticity of LLM-assisted evolutionary search. This suggests that the observed performance gains are not driven by isolated favorable runs but reflect systematic improvements.

To obtain a more fine-grained and statistically powerful comparison, we conduct instance-level paired Wilcoxon signed-rank tests comparing DyCA against two state-of-the-art LES baselines (InstSpecHH and EoH-S). Specifically, we construct paired samples by averaging the performance metric of each testing instance across the independent runs. The non-parametric Wilcoxon test is then applied to these paired instance-level mean scores to evaluate whether the median difference between algorithms is statistically significant. This allows us to assess performance differences across a large number of paired observations, providing strong statistical evidence at the instance level.

\begin{table}[h]
\centering
\caption{Statistical significance of DyCA compared to the strongest baselines (InstSpecHH and EoH-S) using the instance-level paired Wilcoxon signed-rank test. * denotes $p < 0.05$, ** denotes $p < 0.01$, and *** denotes $p < 0.001$.}
\label{tab:wilcoxon}
\begin{tabular}{lll}
\toprule
\textbf{Task} & \textbf{vs. InstSpecHH ($p$-value)} & \textbf{vs. EoH-S ($p$-value)} \\
\midrule
\textbf{CVRP} & $9.6 \times 10^{-9}$ (***) & 0.0002 (***) \\
\textbf{TSP}  & 0.0147 (*) & $4.8 \times 10^{-6}$ (***) \\
\textbf{OBP}  & 0.0001 (***) & 0.0003 (***) \\
\textbf{LLC}  & 0.0122 (*) & 0.0006 (***) \\
\bottomrule
\end{tabular}
\end{table}

Table~\ref{tab:wilcoxon} shows that DyCA achieves statistically significant improvements over both baselines across all tasks, with most comparisons reaching $p < 0.01$ or even $p < 0.001$. These results indicate that DyCA consistently outperforms strong baselines across diverse problem instances rather than benefiting from specific subsets.

Taken together, the run-level analysis characterizes the stability of performance under stochastic optimization, while the instance-level statistical tests demonstrate consistent improvements across heterogeneous instances. The agreement between these two perspectives supports the robustness of DyCA, indicating that its performance gains are both stable across runs and systematic across problem instances.

Overall, these results provide empirical evidence that the advantages reported in the main paper are statistically meaningful and reproducible under independent runs, despite the inherent stochasticity of LLM-assisted evolutionary search.

\subsection{Quantitative analysis of the performance distribution across heterogeneous instances}
\label{appendix_tail_analysis}

To provide an intuitive and fine-grained view of the superiority of DyCA, we conduct a distributional analysis of the designed algorithm portfolios across the testing instances. Figure~\ref{fig:tail_violin} visualizes the performance distributions of all methods using violin plots augmented with embedded box plots. Each distribution is computed over the testing instances, where the instance-level performance is averaged across three independent runs. Furthermore, Tables~\ref{tab:tail} and \ref{tab:head} present the precise quantitative performance on the worst 10\% (\textit{Tail}) and best 10\% (\textit{Head}) of instances, respectively, ensuring a fair comparison of worst- and best-case behavior. The head and tail instances are determined separately for each method based on its own performance ranking. For CVRP, TSP, and OBP, lower values indicate better performance, while higher values are preferred for the LLC task.

\begin{figure}[h!]
  \begin{center}
    \centerline{\includegraphics[width=\columnwidth]{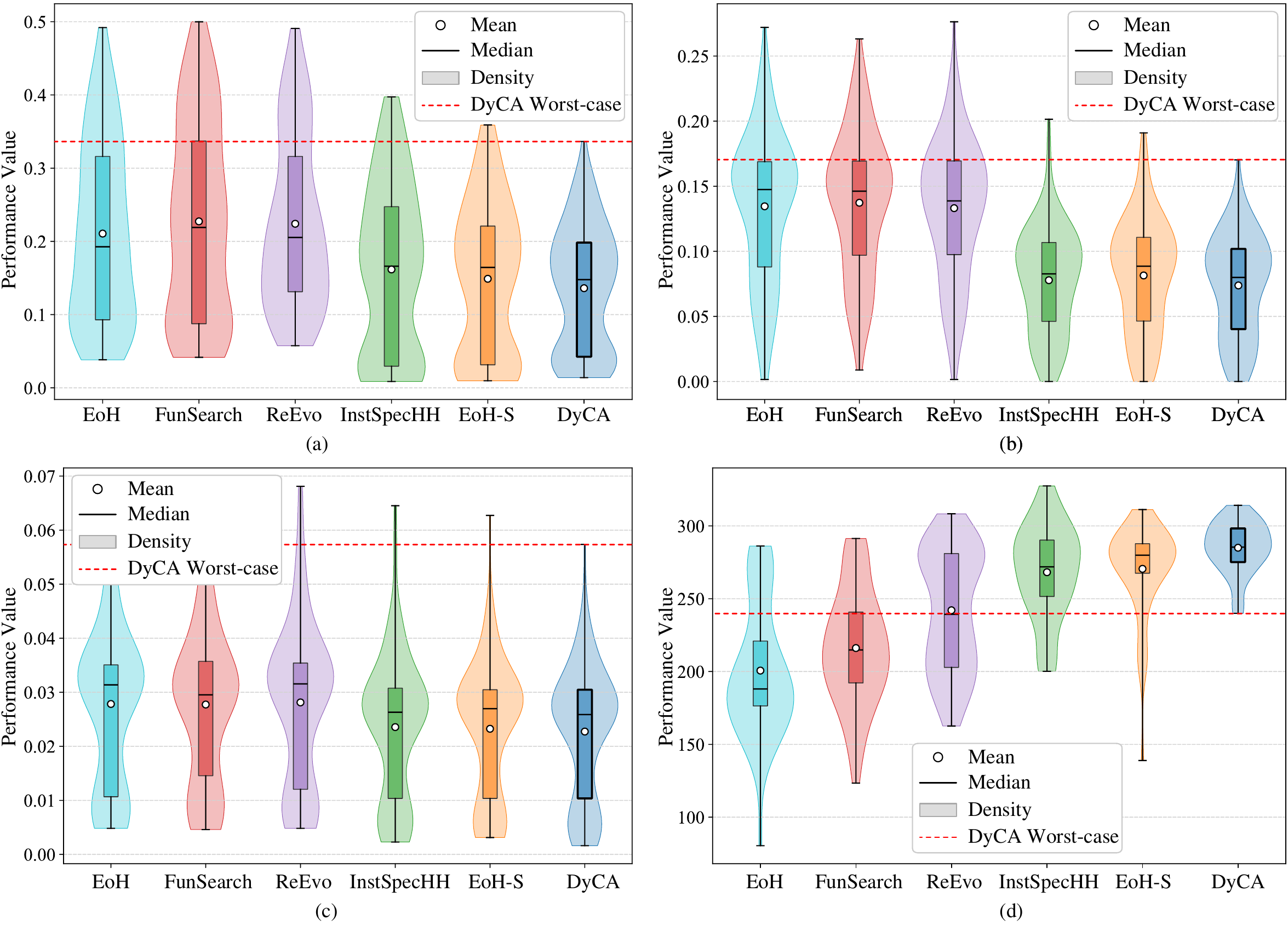}}
    \caption{Performance distributions across testing instances visualized using violin plots with embedded box plots. Panels (a), (b), (c), and (d) correspond to CVRP, TSP, OBP, and LLC, respectively. White circles denote means, black lines denote medians, and shaded regions represent kernel density estimates. The dashed red line indicates the worst-case performance achieved by DyCA. Results highlight DyCA’s improved tail robustness while preserving strong central and head performance.
    }
    \label{fig:tail_violin}
  \end{center}
  \vskip -0.2in
\end{figure}

\begin{table}[h]
\centering
\caption{Comparison of performance on the worst 10\% of instances (Tail).}
\label{tab:tail}
\begin{tabular*}{\linewidth}{l@{\extracolsep{\fill}}cccc}
\toprule
\textbf{Method}     & \textbf{CVRP ($\downarrow$)} & \textbf{TSP ($\downarrow$)} & \textbf{OBP ($\downarrow$)} & \textbf{LLC ($\uparrow$)} \\ \midrule
EoH         & 0.4561 & 0.2324 & 0.0553 & 91.92  \\
FunSearch   & 0.4884 & 0.2503 & 0.0548 & 88.19  \\
ReEvo       & 0.4548 & 0.2348 & 0.0555 & 158.71 \\
EoH-S       & \underline{0.3332} & 0.1588 & \underline{0.0431} & 167.01 \\
InstSpecHH  & 0.3642 & \underline{0.1552} & 0.0469 & \underline{179.96} \\
\textbf{DyCA} & \textbf{0.2890} & \textbf{0.1406} & \textbf{0.0421} & \textbf{220.55} \\ 
\bottomrule 
\end{tabular*}
\end{table}

\begin{table}[h]
\centering
\caption{Comparison of performance on the best 10\% of instances (Head).}
\label{tab:head}
\begin{tabular*}{\linewidth}{l@{\extracolsep{\fill}}cccc}
\toprule
\textbf{Method}     & \textbf{CVRP ($\downarrow$)} & \textbf{TSP ($\downarrow$)} & \textbf{OBP ($\downarrow$)} & \textbf{LLC ($\uparrow$)} \\ \midrule
EoH         & 0.0507 & 0.0263 & 0.0057 & 299.69  \\
FunSearch   & 0.0493 & 0.0271 & 0.0057 & 295.97  \\
ReEvo       & 0.0739 & 0.0295 & 0.0058 & 306.25 \\
EoH-S       & \textbf{0.0144} & \textbf{0.0053} & \underline{0.0036} & 309.70 \\
InstSpecHH  & 0.0164 & 0.0059 & \underline{0.0036} & \textbf{318.71} \\
\textbf{DyCA} & \underline{0.0219} & \underline{0.0054} & \textbf{0.0031} & \underline{316.45} \\ 
\bottomrule 
\end{tabular*}
\end{table}

\paragraph{Tail Robustness.}
A central focus of this analysis is tail robustness, which characterizes worst-case and hard-instance behavior and is particularly critical in heterogeneous instance spaces. The dashed red line in each subplot of Figure~\ref{fig:tail_violin} denotes the worst-case performance achieved by DyCA, serving as a visual reference threshold for robustness. Across all tasks, DyCA exhibits a substantially improved lower tail (or upper tail for LLC) compared to all baseline methods. Notably, DyCA’s performance distributions are consistently shorter and more concentrated in the adverse tail regions, indicating a significant reduction in catastrophic failures and variance on challenging instances. This effect is especially pronounced in the LLC task, where DyCA demonstrates remarkable stability despite the high variability in control difficulty.

\paragraph{Head Performance.}
Importantly, this enhanced tail robustness does not come at the expense of head performance. DyCA simultaneously maintains strong central tendencies, with both its mean and median matching or exceeding those of the strongest competing methods. As detailed in Table~\ref{tab:head}, which reports performance on the best 10\% of testing instances, DyCA consistently ranks among the top two methods across all tasks. The gains achieved in the tail substantially outweigh the minor regressions observed in the head, yielding a highly favorable trade-off. For example, on the CVRP task, DyCA incurs a negligible head performance degradation of 0.0075 relative to EoH-S (0.0219 vs. 0.0144 in Table~\ref{tab:head}), while achieving a substantial tail improvement of 0.0442 (0.3332 vs. 0.2890 in Table~\ref{tab:tail}). This demonstrates that DyCA achieves a favorable balance between robustness and head performance, rather than simply trading average solution quality for conservative worst-case behavior.

Overall, these distributional results corroborate the evaluations presented in the main paper. They demonstrate that DyCA not only improves expected performance but also reshapes the entire performance distribution by compressing adverse tails, thereby yielding more reliable and stable outcomes across heterogeneous instances.


\subsection{Semantic structure discovery and convergence dynamics of B-DISA}
\label{appendix:bdisa_insights}

To better understand the behavior of B-DISA, we conduct an in-depth analysis of its final clustering outcomes and recalibration dynamics on the OBP and CVRP tasks. Beyond demonstrating performance gains, this analysis examines whether B-DISA discovers instance partitions with meaningful structure, and how such structure emerges progressively throughout the algorithm design process.

\paragraph{Final clustering results of B-DISA}
We analyze the clustering results produced by B-DISA after DyCA’s algorithm design process is completed (i.e., after 2{,}000 LLM queries). Based on the dataset descriptions provided in Appendix~\ref{appendix_tasks}, we summarize the final instance partitions identified by B-DISA and their corresponding semantic interpretations in Table~\ref{tab:B-DISA_results}. 

\begin{table}[h]
\centering
\caption{Final B-DISA clustering results and their semantic interpretations.
The reported partitions summarize consistent qualitative patterns observed across three independent DyCA runs.}
\label{tab:B-DISA_results}
\resizebox{\linewidth}{!}{
\begin{tabular}{cccc}
\toprule
\multirow{2}{*}{Task} & \multicolumn{2}{c}{Final B-DISA Clusters} & \multirow{2}{*}{Semantic Interpretation} \\
 & Cluster ID & Training Instance IDs &  \\
\midrule
\multirow{4}{*}{OBP}
 & 1 & 0--80   & \textit{Uniform} item size distribution \\ 
 & 2  & 80--120 & \textit{Weibull} item size distribution (short--medium sequences)\\
 & 3  &  120--160  & \textit{Weibull} item size distribution (long sequences) \\ 
 & 4 & 160--240 & \textit{Bi-modal} item size distribution \\
\midrule
\multirow{5}{*}{CVRP}
 & 1 & 0--60 \& 160--180 & Clustered customers (all scales) and small-scale uniform \\
 & 2 & 180--220 & Large-scale uniform customer distribution \\
 & 3 & 80--160 & Heavy-demand instances (all scales) \\
 & 4 & 60--80 & Clustered customers under high vehicle capacity \\
 & 5 & 220--240 & Uniform customers under high vehicle capacity \\
\bottomrule
\end{tabular}
}
\end{table}

Several notable observations can be drawn from these results. 
\begin{itemize}
\item \textbf{OBP.} B-DISA partitions the 240 instances primarily according to item size distributions, while further uncovering secondary structure within the Weibull regime. Overall, these partitions are consistent with expert knowledge in bin packing, where item size characteristics are widely regarded as the dominant factor governing instance difficulty, and variations in sequence length typically play a secondary role~\cite{falkenauer1996hybrid, gent1998analysis, schwerin1997bin}.

\item \textbf{CVRP.} B-DISA roughly separates the 240 training instances into five clusters, capturing multiple sources of instance heterogeneity. In particular, it (i) explicitly isolates heavy-demand instances as a distinct group, (ii) differentiates clustered and uniform customer distributions under high vehicle capacities, and (iii) further distinguishes uniform instances by problem scale. All of these factors are well known to substantially influence routing difficulty in practice~\cite{solomon1987algorithms, uchoa2017new}.

\end{itemize}

Crucially, these structured and interpretable partitions are obtained without any prior knowledge of instance distributions or handcrafted features. Instead, B-DISA relies entirely on accumulated evaluation data, analyzing instance response patterns to select anchor algorithms. This provides strong empirical evidence that B-DISA can uncover latent instance heterogeneity with genuine semantic meaning, rather than producing arbitrary or purely statistical clusters.

More broadly, these results suggest that B-DISA is capable of revealing instance heterogeneity \textbf{from scratch}. In domains where expert knowledge is limited or unavailable, the final clustering produced by DyCA can serve as a valuable analytical tool, offering insights into which instance characteristics most strongly influence algorithmic performance. In this sense, B-DISA not only supports specialized algorithm design but also has the potential to assist practitioners in constructing or refining domain-specific understanding for previously underexplored problem settings.

\paragraph{Convergence dynamics of B-DISA}
We further investigate how B-DISA converges from an initially uninformative partition to the final clustering described above. To quantify this process, we employ two standard clustering similarity metrics: Adjusted Rand Index (ARI)~\cite{santos2009use} and Normalized Mutual Information (NMI)~\cite{estevez2009normalized}.

ARI measures the agreement between two clustering results and ranges from $-1$ to $1$, where $1$ indicates identical partitions. Values above $0.7$ generally suggest strong similarity. NMI also measures clustering similarity but is less sensitive to the number of clusters, with values ranging from $0$ to $1$.

Using these metrics, we track four quantities throughout the recalibration process:
\begin{itemize}[itemsep=3pt, topsep=0pt, parsep=0pt]
    \item \textbf{Step-ARI}: the ARI between the clustering results of two consecutive recalibration steps.
    \item \textbf{Step-NMI}: the NMI between the clustering results of two consecutive recalibration steps.
    \item \textbf{Ref-ARI}: the ARI between the current clustering and the final clustering (used as a reference).
    \item \textbf{Ref-NMI}: the NMI between the current clustering and the final clustering.
\end{itemize}

\begin{figure}[h]
  \begin{center}
    \centerline{\includegraphics[width=\textwidth]{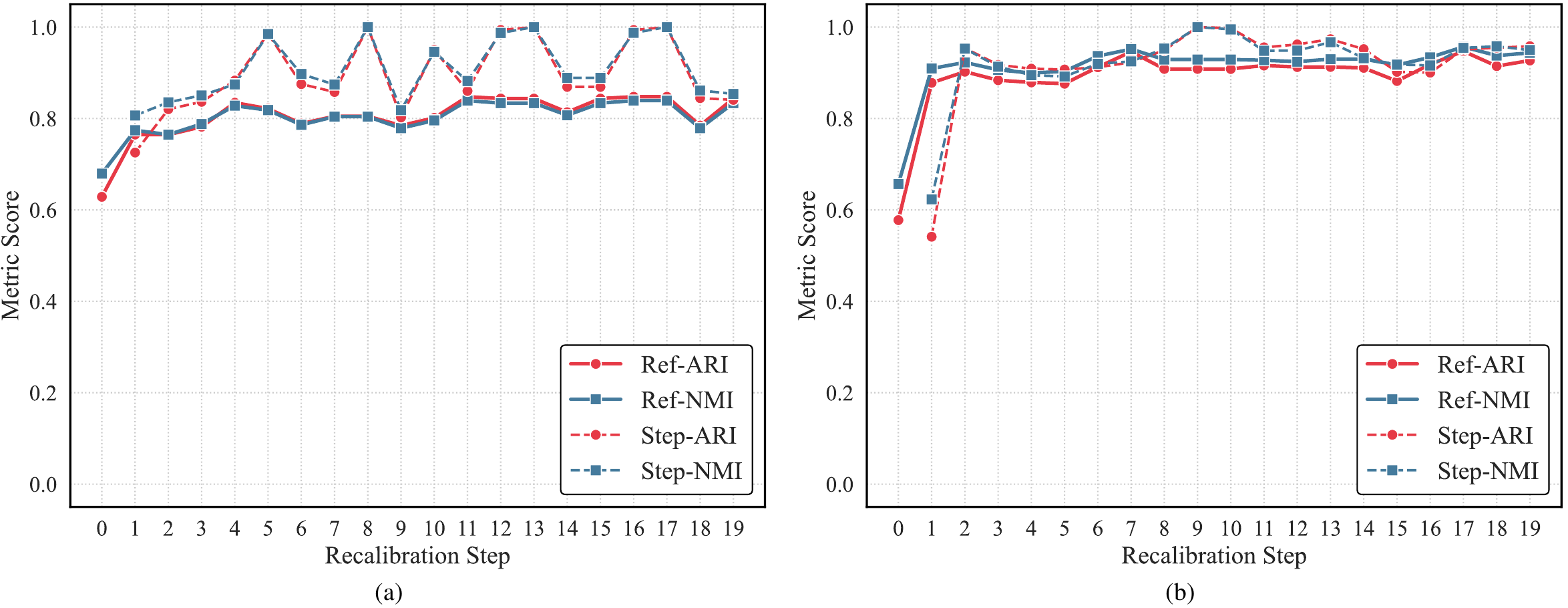}}
    \caption{Convergence behavior of B-DISA during the recalibration process on (a) CVRP and (b) OBP.}
    \label{fig:bdisa_convergence}
  \end{center}
  \vskip -0.2in
\end{figure}

Step-ARI and Step-NMI characterize the self-adjustment behavior of B-DISA. As shown in Figure~\ref{fig:bdisa_convergence}, both metrics fluctuate consistently between $0.8$ and $1.0$ after the first two recalibration rounds and do not exhibit prolonged stagnation at $1.0$ in subsequent iterations. This pattern indicates that B-DISA reaches a regime of local stability without premature freezing: while the overall partition structure remains stable, newly discovered anchor algorithms continue to induce subtle yet meaningful refinements in instance relationships. Such behavior suggests that B-DISA remains sensitive to newly acquired algorithmic information, rather than collapsing into a static clustering early in the design process. These observations empirically support our claim that B-DISA progressively reveals instance heterogeneity through iterative interaction with the evolving algorithm portfolio.

Ref-ARI and Ref-NMI provide a global view of convergence. Both metrics increase rapidly, reaching values close to $0.8$ after the second recalibration, and then remain stable or improve slightly thereafter. This indicates that anchor algorithms discovered within the first $\sim$200 LLM queries already capture the dominant dimensions of instance separability, enabling B-DISA to form a near-final partition at an early stage and thereby allocating more budget to subsequent specialized algorithm design. Later recalibration steps primarily refine cluster boundaries rather than altering the global partition structure. This behavior highlights both (i) the efficiency with which DyCA generates informative anchor algorithms and (ii) the convergence stability of B-DISA as a co-evolving component of the algorithm design process.

Taken together, these results demonstrate that B-DISA does not converge in a post-hoc or static manner. Instead, its clustering structure emerges jointly with the evolution of the algorithm portfolio, exhibiting fast global convergence, continuous local refinement, and strong robustness throughout the design process. This co-evolutionary behavior underpins DyCA’s ability to support effective and stable specialization in heterogeneous settings.

\subsection{Extended ablation analysis of DyCA Components}
\label{add_ablation_study}

This section provides a more fine-grained ablation analysis that complements the results reported in the main paper. While the ablation study in the main text focuses on the core components of DyCA, we further investigate additional design choices and offer deeper insights into the effectiveness of B-DISA for specialized algorithm design.

\begin{table}[h]
\centering
\caption{Ablation Studies on CVRP.}
\label{tab:ablation_appendix}
\begin{tabular}{cccc}
\toprule
\textbf{No.} & \textbf{Setting}        & \textbf{Mean}    & \textbf{Degradation} \\ \midrule
1            & DyCA (Full)                   & \textbf{0.1361}    & -    \\ \midrule
2            & No Clustering           & 0.1530          & 12.46 \%            \\
3            & Random Clustering       & 0.1495          & 9.89 \%              \\
4            & Feature Clustering (Oracle)      & 0.1366          & 0.35 \%              \\ \midrule
5            & w/o Weighting Mechanism & 0.1436 & 5.55 \%              \\
6            & w/o Adaptive Allocation & 0.1459          & 7.25 \%              \\
7            & w/o Specialist Evolution          & 0.1470          & 8.06 \%              \\
8            & w/o Knowledge Sharing     & 0.1434           & 5.35 \%           \\ \bottomrule        
\end{tabular}
\end{table}

\paragraph{Further evidence for the effectiveness of B-DISA.}
Beyond the analysis in the main paper, Variants No.~3 (Random Clustering) and No.~4 (Feature Clustering, Oracle) provide additional insight into the critical role of B-DISA in enabling effective specialization.

Random clustering (No.~3) represents an extreme case of misaligned instance space analysis, where instance groupings bear no semantic relationship to algorithmic behavior. Compared with the oracle setting (No.~4), the only difference lies in the quality of the features used for clustering, while the downstream evolutionary mechanisms remain identical. The substantial performance gap between these two variants indicates that specialization mechanisms are highly sensitive to the quality of instance representations, as cluster assignments directly guide subsequent specialist evolution.

From this perspective, DyCA achieves performance comparable to (and even slightly exceeding) that of the oracle setting, despite relying solely on anchor-induced response vectors without any expert-crafted instance features. This result provides strong empirical evidence that anchor-induced responses constitute a highly informative instance representation, capable of discovering the latent algorithmic structure of the instance space. Consequently, B-DISA enables effective specialization in a fully task-agnostic manner.

\paragraph{Efficacy of bias-aware specialist design.}
Variant No.~6 (w/o Adaptive Allocation) replaces the adaptive resource allocation strategy with a uniform allocation across pools. The observed $7.25\%$ degradation suggests that dynamically allocating computational resources based on evolutionary improvement signals is substantially more effective than static, uniform allocation. This result highlights the importance of bias-aware control in managing heterogeneous specialist pools.

\paragraph{Importance of knowledge sharing.}
Variant No.~8 disables crossover-based knowledge sharing between the complementary pool and specialist pools. The resulting $5.35\%$ performance degradation indicates that knowledge sharing plays a critical role in transferring high-quality algorithmic components across pools. This mechanism helps prevent premature stagnation of individual specialists and contributes to a more robust and powerful algorithm portfolio.

\subsection{Detailed results on public benchmarks}
\label{appendix_benchmarks}

We conduct a comprehensive evaluation of DyCA on widely used public benchmarks and compare it against state-of-the-art LES baselines. These benchmarks exhibit substantial structural and distributional heterogeneity, and thus provide rigorous testbeds for assessing generalization beyond synthetic training distributions. The benchmark datasets considered in this study are summarized below.

\begin{itemize}
    \item \textbf{BPPLib.} We select representative benchmark sets from BPPLib~\cite{delorme2018bpplib}, as summarized in Table~\ref{bpplib_bench}. This collection comprises over 700 instances, with the number of items ranging from 100 to 1{,}002. To ensure consistency across instances, all bin capacities are normalized to 100.

    \item \textbf{TSPLib.} We evaluate 49 commonly used symmetric Euclidean instances from TSPLib~\cite{reinelt1991tsplib}, with problem sizes ranging from 52 to 1{,}000 cities. For consistent evaluation, all coordinates are normalized to the range $[0,1]^2$ by uniformly scaling both dimensions according to the maximum spatial extent, while preserving the original aspect ratio:
    \[
    \text{Scaling factor} = \max(x_{\max} - x_{\min},\; y_{\max} - y_{\min}).
    \]

    \item \textbf{CVRPLib.} We consider seven widely used benchmark sets (A, B, E, F, M, P, and X) from CVRPLib~\cite{uchoa2017new}. The characteristics of these sets are summarized in Table~\ref{cvrp_bench}. Similar to TSPLib, all coordinates are normalized to $[0,1]^2$ using the maximum spatial extent to maintain scale consistency.
\end{itemize}

\begin{table}[htbp]
  \centering
  \caption{Summary of BPPLib benchmark sets.}
  \label{bpplib_bench}
  \begin{tabular}{cccc}
    \toprule
    Benchmark Set & Number of Instances & Capacity & Number of Items \\
    \midrule
    Schwerin\_1~\cite{scholl1997bison} & 100 & 1k & 100 \\
    Schwerin\_2~\cite{scholl1997bison} & 100 & 1k & 120 \\
    AugmentedIRUP~\cite{delorme2016bin} & 250 & \{2.5K--80K\} & \{201--1002\} \\
    AugmentedNonIRUP~\cite{delorme2016bin} & 250 & \{2.5K--80K\} & \{201--1002\} \\
    \bottomrule
  \end{tabular}
\end{table}

\begin{table}[htbp]
  \centering
  \caption{Summary of CVRPLib benchmark sets.}
  \label{cvrp_bench}
  \begin{tabular}{ccc}
    \toprule
    Benchmark Set & Number of Instances & Instance Size \\
    \midrule
    Set A & 27 & 31--79 \\
    Set B & 23 & 30--77 \\
    Set E & 11 & 22--101 \\
    Set F & 3 & 44--134 \\
    Set M & 5 & 100--199 \\
    Set P & 23 & 15--100 \\
    Set X & 43 & 100--300 \\
    \bottomrule
  \end{tabular}
\end{table}

\begin{table}[h!]
\centering
\caption{Detailed results on public benchmarks.}
\label{tab:benchmarks_appendix}
\begin{tabular}{ccccc}
\toprule
\textbf{Benchmarks} & \textbf{EoH} & \textbf{InstSpecHH} & \textbf{EoH-S} & \textbf{DyCA}   \\ \midrule
CVRPLib A           & 0.3402       & 0.2975              & 0.2595         & \textbf{0.2296} \\
CVRPLib B           & 0.3178       & 0.2821              & 0.2204         & \textbf{0.1963} \\
CVRPLib E           & 0.4600       & 0.3150              & 0.2778         & \textbf{0.2376} \\
CVRPLib F           & 0.6470       & 0.4929              & 0.4553         & \textbf{0.4099} \\
CVRPLib M           & 0.4800       & 0.3875              & 0.3467         & \textbf{0.3080} \\
CVRPLib P           & 0.3541       & 0.2542              & 0.2113         & \textbf{0.1663} \\
CVRPLib X           & 0.2122       & 0.2019              & 0.1696         & \textbf{0.1656} \\ \midrule
TSPLib              & 0.1799       & 0.1088              & 0.1127         & \textbf{0.1060} \\ \midrule
BPPLib Sch 1        & 0.1529       & 0.1529              & 0.1528         & \textbf{0.1526} \\
BPPLib Sch 2        & 0.1424       & 0.1422              & 0.1419         & \textbf{0.1413} \\
BPPLib IRUP         & 0.0771       & 0.0738              & 0.0723         & \textbf{0.0694} \\
BPPLib NonIRUP      & 0.0780       & 0.0747              & 0.0732         & \textbf{0.0703} \\ \bottomrule  
\end{tabular}
\end{table}

For all LES methods, the algorithm portfolios trained on synthetic datasets are directly applied to these benchmarks without any fine-tuning or adaptation. This evaluation protocol isolates each method’s intrinsic generalization capability and enables a fair comparison in realistic settings.

As shown in Table~\ref{tab:benchmarks_appendix}, DyCA consistently outperforms all baselines across all evaluated benchmarks. Notably, these improvements are observed across different benchmark families, suggesting that DyCA does not rely on overfitting to specific structural patterns, nor does it sacrifice performance on particular instance types. When transferred directly to unseen real-world instances without adaptation, DyCA demonstrates stronger and more stable generalization than existing LES baselines. Overall, these results provide additional empirical support for the effectiveness and robustness of the mechanisms introduced in DyCA across diverse problem domains.

\section{Algorithmic details of B-DISA}
\label{appendix_bdisa}

The core idea of B-DISA is to characterize problem instances by their \emph{algorithmic response patterns}, rather than by handcrafted or static structural features. Let $\mathcal{A} = {a_1, \dots, a_n}$ denote the archive of $n$ algorithms generated up to the current LES generation. From the evaluation data naturally accumulated during evolution, we construct an algorithm--instance performance matrix $\mathbf{X} \in \mathbb{R}^{M \times n}$, where $X_{m,j}$ denotes the performance of algorithm $a_j$ on instance $i_m$.

Each row $\mathbf{r}_m \in \mathbb{R}^{n}$ of $\mathbf{X}$ represents the \emph{response vector} of instance $i_m$, encoding how this instance reacts to a diverse set of algorithmic strategies. Conversely, each column $\mathbf{p}_j \in \mathbb{R}^{M}$ corresponds to the \emph{performance profile} of algorithm $a_j$ across instances.

By analyzing response vectors, B-DISA groups instances that exhibit similar algorithmic solvability patterns, thereby revealing latent heterogeneity in the instance space and providing a principled basis for specialized algorithm design. However, the dimensionality of response vectors grows continuously with the evolutionary process, rendering direct clustering in $\mathbb{R}^n$ increasingly ineffective.

To address this challenge, B-DISA introduces a compact set of \textbf{Anchor Algorithms} $\mathcal{Z} \subset \mathcal{A}$, which project the high-dimensional matrix $\mathbf{X}$ onto a low-dimensional yet discriminative anchor-induced matrix $\bar{\mathbf{X}} \in \mathbb{R}^{M \times |\mathcal{Z}|}$. Standard clustering methods (e.g., K-Means) are then applied to the rows of $\bar{\mathbf{X}}$ to identify coherent instance clusters that guide specialized evolution. 

In summary, B-DISA exploits $\mathbf{X}$ in two complementary ways:
(i) column-wise analysis to select informative anchor algorithms, and
(ii) row-wise analysis of anchor-induced response vectors to uncover instance structure.

\subsection{Anchor algorithm selection}
\label{appendix_bdisa_anchor_selection}

The goal of anchor selection is to identify a compact subset of algorithms $\mathcal{Z} \subset \mathcal{A}$ that maximally preserves the diversity of instance response patterns.

A key challenge arises from the heterogeneous difficulty of instances. Instances with low difficulty often exhibit large absolute performance ranges, whereas hard instances may show tightly clustered values, with small numerical differences corresponding to qualitatively different outcomes (e.g., infeasible vs.\ feasible). As a result, selection strategies based directly on raw values, variances, or Euclidean distances tend to be dominated by easy instances, obscuring critical structural signals from difficult ones.

To avoid this bias, B-DISA emphasizes relative behavioral tiers rather than absolute numerical magnitudes.

\subsubsection{Preprocessing: natural break discretization}
\label{appendix_bdisa_discretization}

We adopt a Natural Break Discretization scheme that treats each instance row $\mathbf{r}_m$ as a one-dimensional clustering problem. The objective is to map continuous performance values into discrete behavioral tiers defined by significant gaps, thereby normalizing instance contributions when assessing algorithmic discrimination.

Let $\mathbf{v}^{(m)}$ denote the sorted response vector for instance $i_m$. We compute adjacent gaps $g_u = v^{(m)}_{u+1} - v^{(m)}_u$ for $u = 1, \dots, n-1$. A gap is considered a \emph{significant break} if
\begin{equation}
    g_k > \tau_m, \quad
    \tau_m = \max(\eta_{\text{range}} \cdot R_m,\; \eta_{\text{gap}} \cdot \bar{g}_m),
\end{equation}
where $R_m = v^{(m)}_n - v^{(m)}_1$ is the response range for instance $i_m$, and $\bar{g}_m$ is the mean adjacent gap. The threshold $\tau_m$ is designed to be instance-adaptive by combining two complementary criteria.  
The range-based term $\eta_{\text{range}} \cdot R_m$ prevents spurious tier splits when the overall response variation is small, while the gap-based term $\eta_{\text{gap}} \cdot \bar{g}_m$ filters out minor fluctuations that arise from noise or near-ties among algorithms. Taking the maximum of the two ensures that a break is introduced only when it is significant both in absolute and relative terms. In our experiments, we set $\eta_{\text{range}} = 0.15$ and $\eta_{\text{gap}} = 4.0$. These values are chosen to strike a conservative balance between over-segmentation and under-segmentation: they suppress trivial discretization on nearly homogeneous instances while reliably capturing pronounced performance separations when they exist.

Based on the detected breaks, the raw matrix $\mathbf{X}$ is transformed into a discrete matrix $\mathbf{L} \in \mathbb{Z}^{M \times n}$. This discretization ensures that transitions between behavioral tiers contribute comparably across instances, regardless of absolute performance scale, enabling fair and robust comparison of algorithmic discrimination power in subsequent anchor selection operations.

\subsubsection{Anchor selection strategy: greedy search with pruning}
\label{appendix_bdisa_strategy}

Using the discrete matrix $\mathbf{L}$, we define the discriminative power of an algorithm subset $\mathcal{S}$ as the number of distinct response patterns it induces:
\[
\Phi(\mathcal{S}) = \text{the set of unique rows of the sub-matrix defined by } \mathcal{S}.
\]
Our objective is to identify a minimal subset $\mathcal{Z}$ that maximizes $|\Phi(\mathcal{Z})|$. This is achieved via a two-phase procedure.

\paragraph{Phase 1: greedy forward selection.}
Starting from $\mathcal{Z}_0 = \emptyset$, at each step we select
\begin{equation}
    a^* = \arg\max_{a \in \mathcal{A} \setminus \mathcal{Z}_t}
    \left| \Phi(\mathcal{Z}_t \cup \{a\}) \right|.
\end{equation}
Each selected anchor acts as a new probe that further refines instance distinctions. The process terminates when no candidate increases the number of unique patterns or a predefined budget is reached.

\paragraph{Phase 2: backward pruning.}
To eliminate redundancy, we iteratively remove anchors whose exclusion does not reduce discriminative power:
\begin{equation}
    \mathcal{Z} \leftarrow \mathcal{Z} \setminus \{a\}
    \quad \text{if} \quad
    \left| \Phi(\mathcal{Z} \setminus \{a\}) \right|
    =
    \left| \Phi(\mathcal{Z}) \right|.
\end{equation}
This step ensures that the anchor set remains compact without sacrificing resolution.

Once $\mathcal{Z}$ is finalized, clustering is performed on the anchor-induced matrix $\bar{\mathbf{X}}$, yielding instance groups that guide the specialized evolution of the algorithm portfolio.

\subsection{Instance clustering via anchor-induced representations}
\label{appendix_bdisa_clustering}

Given the finalized anchor set $\mathcal{Z}$, each instance $i_m$ is represented by its anchor-induced response vector
\[
\bar{\mathbf{r}}_m = \left( X_{m,z_1}, \dots, X_{m,z_{|\mathcal{Z}|}} \right),
\]
forming the reduced matrix $\bar{\mathbf{X}} \in \mathbb{R}^{M \times |\mathcal{Z}|}$. This representation captures how instances respond to a compact yet maximally discriminative set of algorithmic probes.

\paragraph{Clustering objective.}
The goal of clustering is not to group instances by geometric proximity in the original feature space, but to identify \emph{algorithmically equivalent} instance groups—instances that induce similar relative performance patterns across the anchor algorithms. Such instances are expected to benefit from similar specialized algorithmic strategies, providing a principled basis for downstream specialist evolution.

\paragraph{Automatic determination of the number of clusters.}
The number of clusters $K$ is determined adaptively using an elbow method~\cite{syakur2018integration}. Specifically, we compute the within-cluster sum of squared distances for increasing values of $k$ and detect the point of diminishing returns. This procedure yields a data-driven estimate of $K$ that balances resolution and robustness, avoiding both over-partitioning and under-segmentation of the instance space.


\paragraph{Clustering method.}
We apply $k$-means clustering to the rows of $\bar{\mathbf{X}}$ using Euclidean distance. Although simple, this choice is effective because the anchor selection process explicitly maximizes behavioral discrimination, rendering $\bar{\mathbf{X}}$ low-dimensional, well-conditioned, and suitable for centroid-based partitioning.

\subsection{Dynamic recalibration during the LES process}
\label{appendix_recalibration}

B-DISA is designed to evolve jointly with the LES process. As new algorithms are discovered, novel behavioral distinctions may emerge, necessitating periodic updates of anchor algorithms and instance clusters.

To maintain scalability, we adopt an \textbf{incremental anchor update strategy}. Let $\mathcal{Z}_{t-1}$ be the anchor set obtained in the previous cycle, and let $\mathcal{P}_{\text{new}}$ denote the newly generated candidate algorithms. The effective candidate pool at cycle $t$ is defined as
\begin{equation}
    \mathcal{S}_t = \mathcal{Z}_{t-1} \cup \mathcal{P}_{\text{new}}.
\end{equation}
The greedy selection and pruning procedure described above is then applied solely to $\mathcal{S}_t$.

This design yields three key benefits:
\begin{enumerate}
    \item \textbf{Computational Efficiency:} The candidate space is significantly reduced, ensuring scalability as the algorithm archive grows.
    \item \textbf{Continuity:} Previously selected anchors are retained unless newly discovered algorithms provide strictly stronger or complementary discriminative capability.
    \item \textbf{Monotonic Discriminative Power:} Since $\mathcal{Z}_{t-1} \subset \mathcal{S}_t$, the resulting anchor set $\mathcal{Z}_t$ satisfies
    $|\Phi(\mathcal{Z}_t)| \ge |\Phi(\mathcal{Z}_{t-1})|$, guaranteeing non-decreasing resolution of instance heterogeneity and stable refinement of instance clusters.
\end{enumerate}

\section{Details of bias-aware search effort allocation}
\label{appendix_specialist_pool}

In DyCA, specialist algorithm design is carried out by maintaining multiple specialist pools
$\{\mathcal{C}_1, \dots, \mathcal{C}_K\}$, each associated with an instance cluster identified by B-DISA.
A critical design question is how to allocate the limited LLM inference budget among these pools.
Uniform allocation is inefficient, as different clusters exhibit varying levels of difficulty and different degrees of progress toward their respective sub-objectives.

To address this challenge, DyCA adopts a \textbf{Bias-Aware Search Effort Allocation} strategy equipped with an explicit \textbf{Circuit Breaker} mechanism.
The core principle is to bias computational resources toward clusters that are underperforming or inherently difficult, while preventing excessive allocation to clusters that fail to respond to further optimization.

\subsection{Estimation of cluster progress}
\label{app_cluster_pro_estimation}

Let $T_k$ denote the target performance for cluster $k$, such as a theoretical upper bound or the best-known achievable score for the corresponding sub-objective.
For the $k$-th specialist pool, let $T^{(k)}_{\text{best}}$ denote the best performance historically achieved on instances belonging to cluster $k$.

We first define the raw performance gap
\begin{equation}
    \Delta_k = \max\left(0,\, T_k - T^{(k)}_{\text{best}}\right),
\end{equation}
which quantifies the remaining improvement potential of cluster $k$.
Intuitively, a larger $\Delta_k$ indicates that the cluster is further from its sub-objective and thus may benefit more from additional search effort.

However, allocating resources solely based on $\Delta_k$ is insufficient.
Some clusters may be intrinsically difficult or poorly aligned with the current algorithmic search space, and persistently allocating resources to such clusters can lead to stagnation and inefficient budget usage.
To mitigate this issue, DyCA introduces a stagnation-aware decay mechanism.

Let $\eta_k$ denote the number of consecutive generations during which pool $\mathcal{C}_k$ fails to improve its best performance.
The effective allocation weight $W_k$ is then defined as
\begin{equation}
    W_k =
    \begin{cases}
        +\infty, & \text{if } T^{(k)}_{\text{best}} = -\infty \quad \text{(cold start)}, \\
        \Delta_k \cdot \max\!\left(\gamma_{\min},\, \alpha^{\eta_k}\right), & \text{otherwise},
    \end{cases}
\end{equation}
where $\alpha \in (0,1)$ is a decay rate and $\gamma_{\min} > 0$ is a minimum weight floor.

The decay factor $\alpha^{\eta_k}$ progressively down-weights clusters that fail to exhibit improvement, acting as a circuit breaker that prevents unproductive pools from monopolizing the inference budget.
The lower bound $\gamma_{\min}$ ensures that each cluster retains a minimal level of exploration, avoiding premature exclusion.
The cold-start condition guarantees that newly formed or previously unexplored clusters are immediately prioritized.

\subsection{Probabilistic allocation of search effort}
\label{app_how_to_allocate}

Given the effective weights $\{W_k\}_{k=1}^{K}$, DyCA allocates search effort via weighted sampling.
The probability of selecting specialist pool $\mathcal{C}_k$ for the next LLM-assisted evolutionary step is defined as
\begin{equation}
    P(k) = \frac{W_k}{\sum_{j=1}^{K} W_j}.
\end{equation}

This probabilistic formulation yields a dynamic and self-regulating allocation behavior.
Clusters with large remaining improvement potential naturally attract more search effort, while clusters exhibiting prolonged stagnation are gradually deprioritized.
As a result, DyCA consistently allocates computational resources to the most underserved regions of the instance space, thereby improving the efficiency and robustness of specialized algorithm evolution.

\section{Task descriptions and heterogeneous instance set construction}
\label{appendix_tasks}

We evaluate DyCA across four representative automated algorithm design tasks spanning routing, packing, and online decision-making problems. Each task is constructed with explicitly heterogeneous instance distributions to assess the framework’s ability to discover latent instance structures and to support bias-aware specialized evolution.

\subsection{Traveling Salesman Problem (TSP)}
\label{appendix_task_tsp}

\textbf{Problem definition.} 
Let $G = (V, E)$ be a complete graph, where $V = \{v_1, \dots, v_n\}$ denotes a set of $n$ cities with coordinates $\mathbf{x}_i \in [0,1]^2$. Each edge $(i,j) \in E$ is associated with a Euclidean distance cost $c_{ij} = \|\mathbf{x}_i - \mathbf{x}_j\|_2$. The objective is to find a Hamiltonian cycle $\pi = (\pi_1, \dots, \pi_n, \pi_1)$ that minimizes the total tour length:
\begin{equation}
    L_{\text{TSP}} = \sum_{k=1}^{n-1} c_{\pi_k \pi_{k+1}} + c_{\pi_n \pi_1}.
\end{equation}

\textbf{Designed algorithm.} 
Rather than solving TSP directly, DyCA aims to design a constructive heuristic. The target algorithm incrementally builds a tour by selecting the next city at each step. At every decision point, the heuristic receives as input the current city, the set of unvisited cities, and the distance matrix, and outputs the index of the next city to visit. Algorithm performance is evaluated by the relative gap with respect to high-quality reference solutions produced by the LKH-3 solver~\cite{helsgaun2017extension}.

\begin{figure}[h]
  \begin{center}
    \centerline{\includegraphics[width=0.9\textwidth]{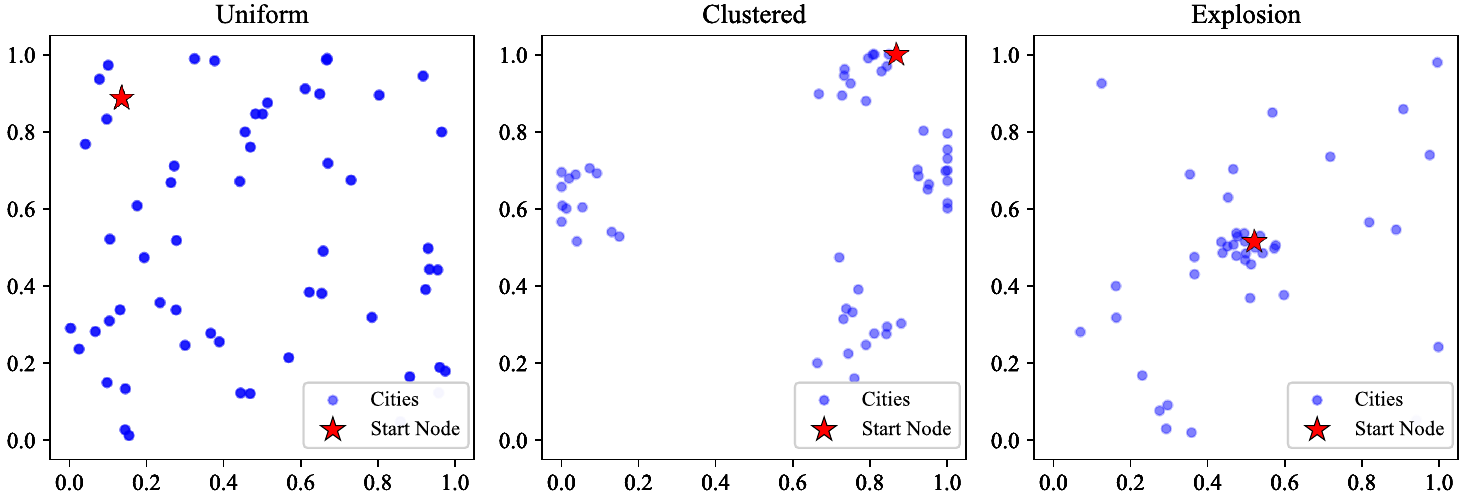}}
    \caption{Visualization of the three spatial distributions used in the TSP task. \textbf{Left:} Uniform distribution. \textbf{Middle:} Clustered distribution with multiple dense regions. \textbf{Right:} Explosion distribution exhibiting a dense core and sparse periphery. These distributions impose qualitatively different challenges on constructive routing heuristics.}
    \label{fig:TSP_example}
  \end{center}
  \vskip -0.2in
\end{figure}

\textbf{Instance generation protocol.}
To induce heterogeneous structural characteristics in the instance space, we generate TSP instances from three distinct spatial distributions, as illustrated in Figure~\ref{fig:TSP_example}:
\begin{itemize}[itemsep=3pt, topsep=0pt, parsep=0pt]
    \item \textbf{Uniform.} Cities are sampled uniformly from the unit square $[0,1]^2$. This represents the standard unstructured benchmark commonly used in TSP studies.
    \item \textbf{Clustered.} Cities are generated around $3$ to $6$ randomly selected centroids. Points within each cluster follow a Gaussian distribution with standard deviation $\sigma = 0.07$. This distribution introduces clear sub-regions and tests the algorithm’s ability to exploit local structure while coordinating inter-cluster connections.
    \item \textbf{Explosion.} A hybrid distribution where $50\%$ of the cities are densely concentrated around the center $(0.5, 0.5)$ with $\sigma = 0.05$, while the remaining cities are uniformly distributed over the entire space. This setting creates a strong imbalance between dense and sparse regions, requiring the algorithm to balance local intensification and global exploration.
\end{itemize}

\textbf{Dataset composition.}
By combining four problem scales with three spatial distributions, we construct 12 heterogeneous TSP datasets. For each dataset, 20 instances are used for training and 10 for testing. Detailed specifications of all datasets are summarized in Table~\ref{tab:tsp_dataset}.

\begin{table}[h]
\centering
\caption{Specification of the heterogeneous TSP datasets. Four problem scales are evaluated under three distinct spatial distributions.}
\label{tab:tsp_dataset}
\begin{tabular}{@{}clcccc@{}}
\toprule
\textbf{Group ID} & \textbf{Scale} & \textbf{Distribution} & \textbf{Problem Size ($n$)} & \textbf{Training IDs} & \textbf{Testing IDs} \\ \midrule
1 & \multirow{3}{*}{Small} & Clustered & $10$--$30$ & $[0,20)$ & $[0,10)$ \\
2 &  & Explosion & $10$--$30$ & $[80,100)$ & $[40,50)$ \\
3 &  & Uniform & $10$--$30$ & $[160,180)$ & $[80,90)$ \\ \midrule

4 & \multirow{3}{*}{Medium} & Clustered & $40$--$60$ & $[20,40)$ & $[10,20)$ \\
5 &  & Explosion & $40$--$60$ & $[100,120)$ & $[50,60)$ \\
6 &  & Uniform & $40$--$60$ & $[180,200)$ & $[90,100)$ \\ \midrule

7 & \multirow{3}{*}{Large} & Clustered & $120$--$150$ & $[40,60)$ & $[20,30)$ \\
8 &  & Explosion & $120$--$150$ & $[120,140)$ & $[60,70)$ \\
9 &  & Uniform & $120$--$150$ & $[200,220)$ & $[100,110)$ \\ \midrule

10 & \multirow{3}{*}{Extra Large} & Clustered & $180$--$200$ & $[60,80)$ & $[30,40)$ \\
11 &  & Explosion & $180$--$200$ & $[140,160)$ & $[70,80)$ \\
12 &  & Uniform & $180$--$200$ & $[220,240)$ & $[110,120)$ \\ \bottomrule
\end{tabular}
\end{table}

\subsection{Capacitated Vehicle Routing Problem (CVRP)}
\label{appendix_task_cvrp}

\textbf{Problem definition.} The CVRP generalizes the TSP to a multi-vehicle setting with explicit capacity constraints. Given a depot $v_0$ and a set of customers $\{v_1, \dots, v_n\}$, each customer $v_i$ is associated with a location $\mathbf{x}_i \in [0,1]^2$ and an integer demand $d_i \in \mathbb{Z}^+$, where $d_0 = 0$.
Each vehicle has a fixed capacity $Q \in \mathbb{Z}^+$, and the travel cost between any two locations is defined as the Euclidean distance $c_{ij} = \lVert \mathbf{x}_i - \mathbf{x}_j \rVert_2$.

The objective is to find a set of routes $\mathcal{R} = \{r_1, \dots, r_m\}$ such that each route starts and ends at the depot $v_0$, every customer is served exactly once, and the total demand along each route satisfies the capacity constraint
$\sum_{v_i \in r_k} d_i \le Q$. The goal is to minimize the total travel distance of all routes.

\textbf{Designed algorithm.} As in the TSP task, DyCA aims to design a constructive routing heuristic. The target algorithm incrementally constructs routes by selecting the next customer to serve at each step, conditioned on the current vehicle location, the set of unserved customers, and the remaining vehicle capacity. The performance of the designed algorithms is evaluated using the optimality gap relative to LKH-3.

\textbf{Instance generation protocol.}
To induce diverse logistical scenarios and heterogeneous algorithmic responses, we generate CVRP instances with varying spatial and demand characteristics:
\begin{itemize}[itemsep=3pt, topsep=0pt, parsep=0pt]
    \item \textbf{Uniform.} Customer coordinates are sampled uniformly from $[0,1]^2$, with demands drawn from a uniform distribution $U[1, 9]$. This setting serves as a balanced baseline.
    \item \textbf{Clustered.} Customers are generated around $3$ to $6$ randomly sampled centroids using a Gaussian mixture model with standard deviation $\sigma = 0.06$. This configuration simulates geographically concentrated delivery regions.
    \item \textbf{Heavy Demand.} Customer coordinates are sampled uniformly, while demands are drawn from $U[0.4Q, 0.7Q]$. In this regime, vehicles can typically serve only one or two customers per route, shifting the optimization emphasis from routing efficiency to capacity-aware packing.
\end{itemize}

\textbf{Dataset composition.} By jointly varying the problem scale (number of customers $n$), capacity constraints ($Q$), and instance distributions, we construct a total of 12 CVRP datasets. For each dataset, 20 instances are used for training and 10 instances for testing.
The detailed specifications of all datasets are summarized in Table~\ref{tab:cvrp_dataset}.

\begin{table}[h]
\centering
\caption{Specification of the heterogeneous CVRP datasets. The benchmark covers varying node scales and capacity constraints.}
\label{tab:cvrp_dataset}
\begin{tabular}{@{}clccccc@{}}
\toprule
\textbf{Group ID} & \textbf{Config} & \textbf{Distribution} & \textbf{Size ($n$)} & \textbf{Cap ($Q$)} & \textbf{Training IDs} & \textbf{Testing IDs} \\ \midrule
1 & \multirow{3}{*}{Baseline} & Clustered & $30$--$60$ & $30$--$60$ & $[0,20)$ & $[0,10)$ \\
2 & & Heavy & $30$--$60$ & $30$--$60$ & $[80,100)$ & $[40,50)$ \\
3 & & Uniform & $30$--$60$ & $30$--$60$ & $[160,180)$ & $[80,90)$ \\ \midrule

4 & \multirow{3}{*}{Scale Up} & Clustered & $80$--$120$ & $30$--$60$ & $[20,40)$ & $[10,20)$ \\
5 & & Heavy & $80$--$120$ & $30$--$60$ & $[100,120)$ & $[50,60)$ \\
6 & & Uniform & $80$--$120$ & $30$--$60$ & $[180,200)$ & $[90,100)$ \\ \midrule

7 & \multirow{3}{*}{Large} & Clustered & $180$--$200$ & $30$--$60$ & $[40,60)$ & $[20,30)$ \\
8 & & Heavy & $180$--$200$ & $30$--$60$ & $[120,140)$ & $[60,70)$ \\
9 & & Uniform & $180$--$200$ & $30$--$60$ & $[200,220)$ & $[100,110)$ \\ \midrule

10 & \multirow{3}{*}{High Cap} & Clustered & $80$--$120$ & $90$--$110$ & $[60,80)$ & $[30,40)$ \\
11 & & Heavy & $80$--$120$ & $90$--$110$ & $[140,160)$ & $[70,80)$ \\
12 & & Uniform & $80$--$120$ & $90$--$110$ & $[220,240)$ & $[110,120)$ \\ \bottomrule
\end{tabular}
\end{table}

\subsection{Online Bin Packing (OBP)}
\label{appendix_task_obp}

\textbf{Problem definition.} 
The Online Bin Packing (OBP) problem considers packing a sequence of items into a minimum number of bins, each with a fixed capacity $C \in \mathbb{R}^+$. 
Items arrive sequentially and must be irrevocably assigned to bins upon arrival, without knowledge of future items.

Formally, let $I = (i_1, i_2, \dots, i_n)$ denote a sequence of $n$ items, where each item $i_j$ has a size $s_j \in (0, C]$. 
A bin is a container with capacity $C$, and a packing is feasible if, for every bin $B_k$,
\begin{equation}
    \sum_{i_j \in B_k} s_j \leq C.
\end{equation}
The objective is to minimize the total number of bins used:
\begin{equation}
    K = |\{B_1, B_2, \dots, B_K\}|.
\end{equation}

In the online setting, the following constraints apply: 
(i) items arrive one-by-one in an unknown order; 
(ii) each item must be assigned immediately upon arrival; and 
(iii) previously packed items cannot be moved or reassigned.

\textbf{Designed algorithm.} DyCA aims to design online packing heuristics that compute a priority score over currently open bins for each incoming item, conditioned on the item size and the residual capacities of existing bins. The item is then assigned to the highest-priority bin, or to a new bin if no feasible bin exists. Algorithm performance is evaluated by the excess ratio over the theoretical lower bound on the minimum number of bins required~\cite{martello1990lower}.

\begin{figure}[h]
  \begin{center}
    \centerline{\includegraphics[width=\textwidth]{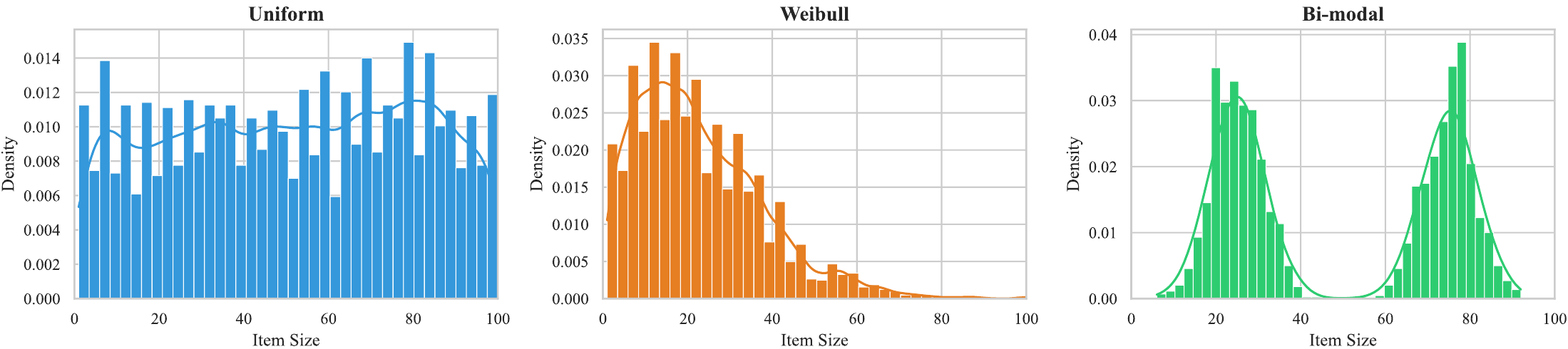}}
    \caption{Illustration of the three item size distributions used in the OBP task: \textbf{Uniform}, \textbf{Weibull}, and \textbf{Bi-modal}. These distributions induce qualitatively different online packing behaviors and decision trade-offs.}
    \label{fig:obp_distribution}
  \end{center}
  \vskip -0.2in
\end{figure}

\textbf{Instance generation protocol.}
To induce heterogeneous online decision-making challenges, we generate OBP instances using three distinct item size distributions with a fixed bin capacity $C = 100$:
\begin{itemize}[itemsep=3pt, topsep=0pt, parsep=0pt]
    \item \textbf{Uniform.} Item sizes are sampled uniformly from $[1, C]$. This serves as a neutral baseline distribution.
    \item \textbf{Weibull.} Item sizes follow a Weibull distribution with shape parameter $k = 1.5$ and scale $\lambda = 0.25C$. This setting mimics long-tailed real-world data, where small items dominate but large items occasionally appear.
    \item \textbf{Bi-modal.} Item sizes are sampled from two distinct modes: small items in $[0.1C, 0.2C]$ and large items in $[0.8C, 0.9C]$. This ``dumbbell'' distribution challenges algorithms to effectively pair complementary items rather than greedily filling bins.
\end{itemize}

Figure~\ref{fig:obp_distribution} visualizes these three item size distributions.

\textbf{Dataset composition.}
The OBP benchmark consists of 12 datasets covering four distinct sequence length regimes, ranging from short to very long horizons. 
Consistent with the TSP and CVRP tasks, each dataset contains 20 training instances and 10 test instances.
Detailed specifications of all datasets are summarized in Table~\ref{tab:obp_dataset}.

\begin{table}[h]
\centering
\caption{Specification of the heterogeneous OBP datasets. The benchmark evaluates performance across varying stream lengths and item size distributions.}
\label{tab:obp_dataset}
\begin{tabular}{@{}clcccc@{}}
\toprule
\textbf{Group ID} & \textbf{Horizon} & \textbf{Distribution} & \textbf{Length ($n$)} & \textbf{Training IDs} & \textbf{Testing IDs} \\ \midrule
1 & \multirow{3}{*}{Short} & Uniform & $200$--$500$ & $[0, 20)$ & $[0, 10)$ \\
2 &  & Weibull & $200$--$500$ & $[80, 100)$ & $[40, 50)$ \\
3 &  & Bi-modal & $200$--$500$ & $[160, 180)$ & $[80, 90)$ \\ \midrule

4 & \multirow{3}{*}{Medium} & Uniform & $800$--$1000$ & $[20, 40)$ & $[10, 20)$ \\
5 &  & Weibull & $800$--$1000$ & $[100, 120)$ & $[50, 60)$ \\
6 &  & Bi-modal & $800$--$1000$ & $[180, 200)$ & $[90, 100)$ \\ \midrule

7 & \multirow{3}{*}{Long} & Uniform & $1200$--$1500$ & $[40, 60)$ & $[20, 30)$ \\
8 &  & Weibull & $1200$--$1500$ & $[120, 140)$ & $[60, 70)$ \\
9 &  & Bi-modal & $1200$--$1500$ & $[200, 220)$ & $[100, 110)$ \\ \midrule

10 & \multirow{3}{*}{Very Long} & Uniform & $1700$--$2000$ & $[60, 80)$ & $[30, 40)$ \\
11 &  & Weibull & $1700$--$2000$ & $[140, 160)$ & $[70, 80)$ \\
12 &  & Bi-modal & $1700$--$2000$ & $[220, 240)$ & $[110, 120)$ \\ \bottomrule
\end{tabular}%
\end{table}

\subsection{Lunar Lander Control (LLC)}
\label{appendix_task_llc}

\textbf{Problem definition.} 
The Lunar Lander Control (LLC) task considers a control problem in which an agent must safely land a spacecraft on a designated landing pad under gravity, using limited thrust actions. The environment is adapted from the standard LunarLander benchmark~\cite{gadgil2020solving}. At each time step $t$, the lander is described by a continuous state vector $\mathbf{s}_t = (x_t, y_t, \dot{x}_t, \dot{y}_t, \theta_t, \dot{\theta}_t)$, which represents the lander’s position, velocity, orientation, and angular velocity. The agent selects control actions corresponding to the main engine thrust and side engine torques, which jointly influence the lander’s translational and rotational dynamics. The objective is to control the lander to a smooth touchdown on the landing pad, minimizing fuel consumption while avoiding crashes or excessive impact velocity.

\textbf{Designed algorithm.} 
DyCA aims to design reactive control policies that map the current state $\mathbf{s}_t$ to continuous control actions.
Rather than learning a neural policy end-to-end, LES searches over programmatic policy representations that compute thrust commands based on interpretable, state-dependent rules. The quality of a candidate controller is evaluated by executing it from a given initial state and measuring the cumulative landing reward, which penalizes crashes, hard landings, and excessive fuel usage while rewarding stable and centered touchdowns. A landing episode with cumulative reward exceeding $200$ is regarded as a safe, efficient, and successful landing~\cite{hu2025mles}.

\textbf{Dataset composition.}
To explicitly introduce heterogeneity in control difficulty, we construct LLC instances by varying the lander's initial state. The training set contains 35 instances covering a wide range of initial velocities, and orientations.
These initial conditions induce qualitatively different uncontrolled descent trajectories, as illustrated in Figure~\ref{fig:LLC_training}. Some instances require highly aggressive and early thrust corrections to prevent divergence, while others admit smoother and more fuel-efficient control strategies. The testing set consists of 50 instances whose initial states are fully covered by the training distribution, as visualized in Figure~\ref{fig:LLC_testing}.

\begin{figure}[h!]
  \begin{center}
    \centerline{\includegraphics[width=0.8\textwidth]{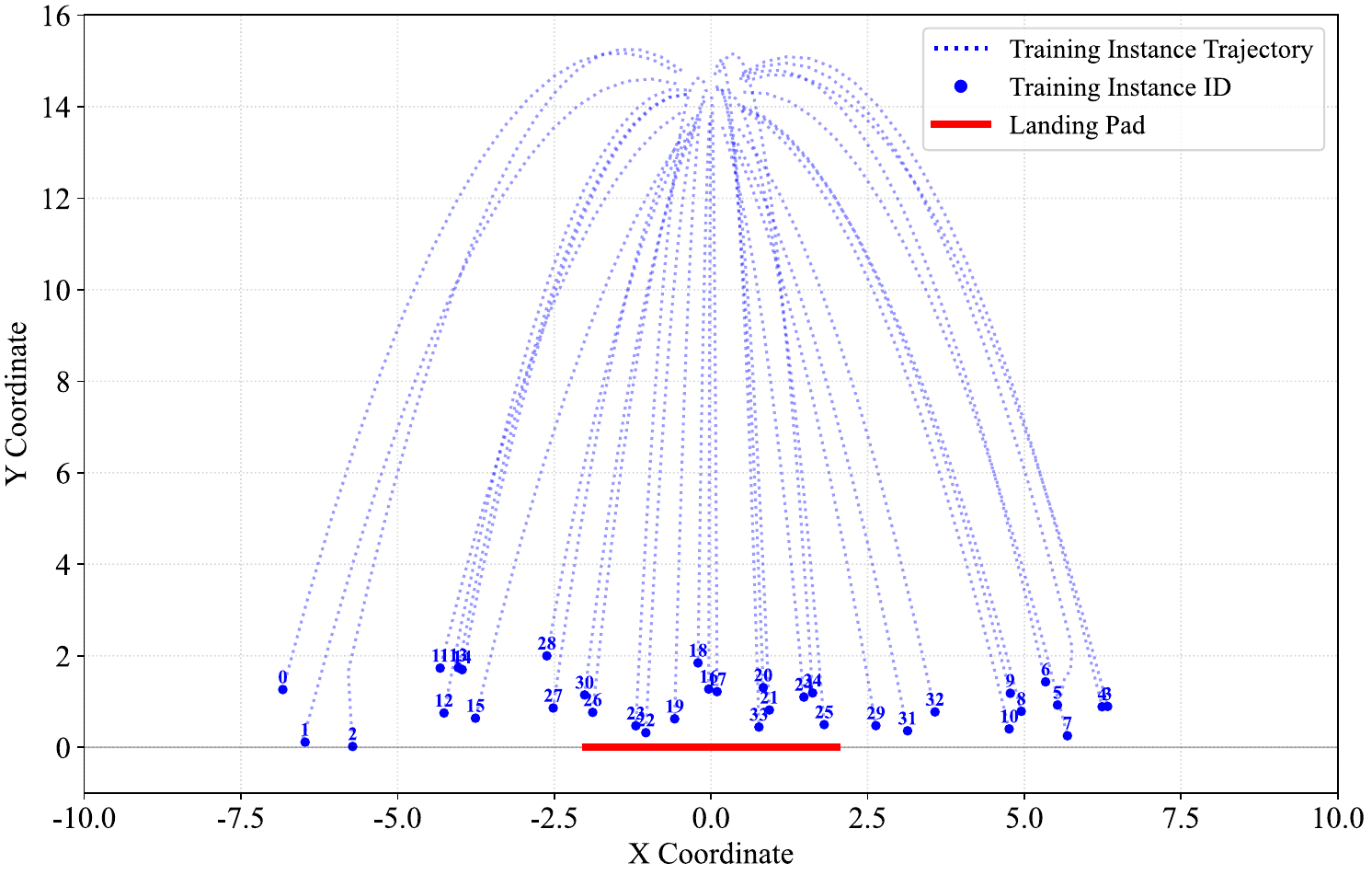}}
    \caption{Uncontrolled descent trajectories of the 35 training instances in the LLC task. Different initial states lead to distinct descent patterns, reflecting varying levels of control difficulty and necessitating control policies with different degrees of aggressiveness.}
    \label{fig:LLC_training}
  \end{center}
  \vskip -0.2in
\end{figure}

\begin{figure}[h!]
  \begin{center}
    \centerline{\includegraphics[width=0.8\textwidth]{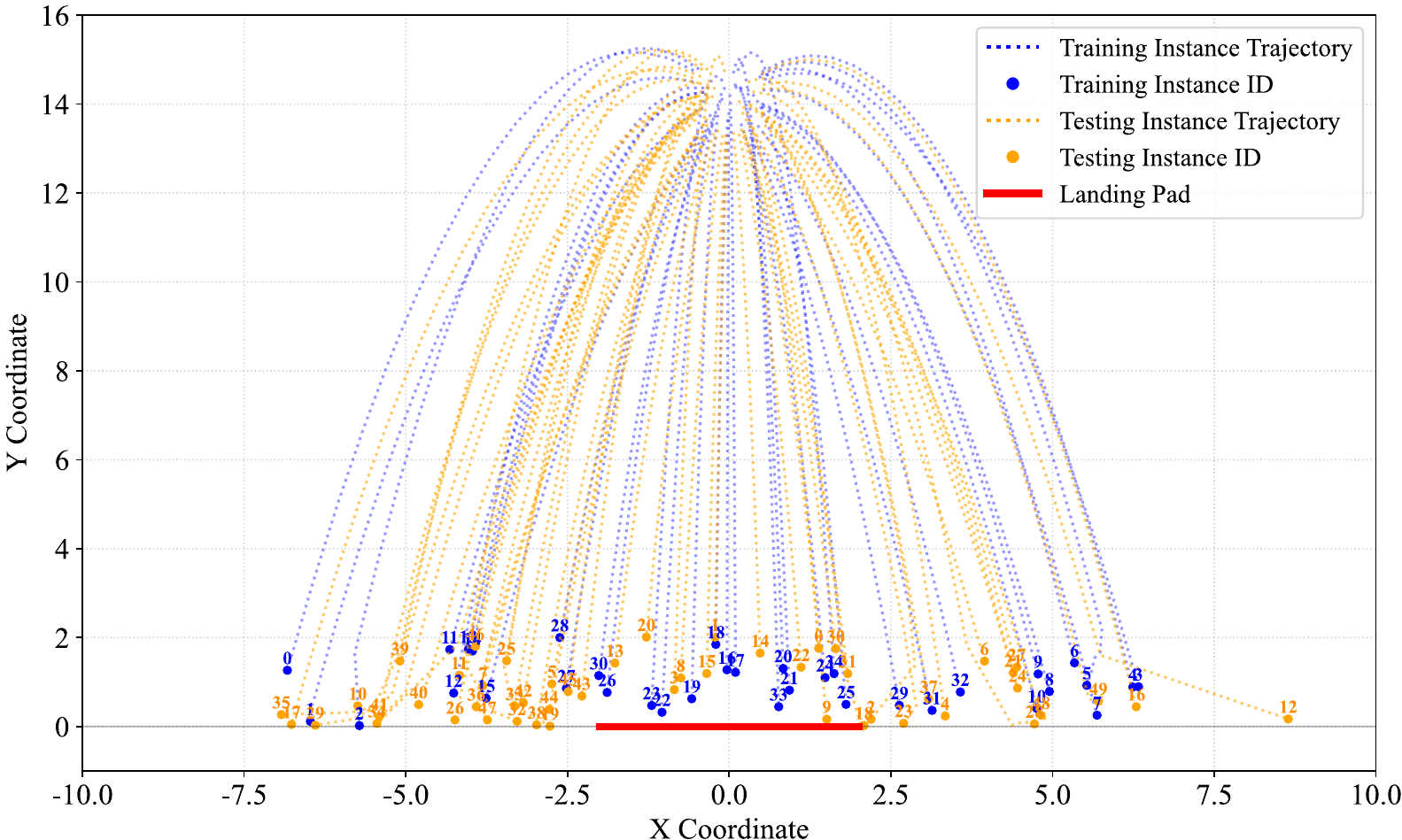}}
    \caption{Initial state distribution of the $50$ testing instances in the LLC task. All testing states are contained within the coverage of the training distribution, ensuring evaluation under interpolative yet heterogeneous control conditions.}
    \label{fig:LLC_testing}
  \end{center}
  \vskip -0.2in
\end{figure}

\section{Theoretical property of weighted CPM}
\label{Theoretical_property_WCPM}

We show that the weighted formulation preserves the key properties required by the greedy selection procedure in CPM.

\paragraph{Reformulation.}
Define the weighted objective over an algorithm pool $P$ as:
\begin{equation}
F_w(P) = \sum_{k=1}^{K} \sum_{i_m \in \mathcal{C}_k} \frac{1}{|\mathcal{C}_k|} \cdot f^*_m(P),
\end{equation}
where $f^*_m(P) = \min_{a \in P} f_m(a)$ (or max depending on formulation). Then the weighted marginal gain can be written as:
\begin{equation}
\Delta \mathrm{CPI}_w(\tilde{a} \mid P) = F_w(P \cup \{\tilde{a}\}) - F_w(P).
\end{equation}

\paragraph{Non-negativity (monotonicity).}
For any instance $i_m$, we have
\[
f^*_m(P \cup \{\tilde{a}\}) \leq f^*_m(P),
\]
which implies that each term in the summation is non-negative after applying the improvement operator:
\[
\max(f^*_m(P) - f_m(\tilde{a}), 0) \geq 0.
\]
Since all weights $\frac{1}{|\mathcal{C}_k|} > 0$, it follows that:
\[
\Delta \mathrm{CPI}_w(\tilde{a} \mid P) \geq 0.
\]
Thus, $F_w(P)$ is monotone.

\paragraph{Additivity.}
The objective $F_w(P)$ is a weighted sum over independent instance-wise contributions:
\[
F_w(P) = \sum_m w_m \cdot f^*_m(P), \quad w_m > 0.
\]
Therefore, the marginal gain decomposes additively across instances:
\[
\Delta \mathrm{CPI}_w(\tilde{a} \mid P) = \sum_m w_m \cdot \Delta_m(\tilde{a} \mid P),
\]
which preserves the additive structure of the original CPM objective.

\paragraph{Submodularity.}
For each instance $i_m$, the function $f^*_m(P)$ is a pointwise minimum over a set, which is known to be a submodular set function. A non-negative weighted sum of submodular functions remains submodular. Therefore, $F_w(P)$ is submodular.

\paragraph{Conclusion.}
Since $F_w(P)$ is monotone and submodular, the standard greedy selection strategy used in CPM retains its theoretical guarantees (e.g., approximation bounds for maximizing a monotone submodular function under cardinality constraints). The weighting only rescales instance contributions without altering these structural properties.

\section{Prompts used in experiments}
\label{all_prompt}

This section documents all prompt templates used in our experiments, including task-specific prompts and prompts for different evolutionary operators.  Our goal is to provide a transparent and reproducible description of the interaction interface between the LLM and the evolutionary search process, while abstracting away task-internal details that are not central to the proposed methodology.

\subsection{Task-specific prompts}
\label{task_prompts}

For each task, we employ a concise task description prompt that specifies the problem setting and the expected form of the algorithmic solution. These prompts serve solely to define the optimization objective and the step-wise decision structure, without injecting any domain-specific heuristics or solution strategies.
The task-specific prompts used for TSP, CVRP, OBP, and LLC are shown in Figures~\ref{fig:TSP_taskdescription}--\ref{fig:Lunar_taskdescription}, respectively.

\begin{figure}[h!]
\centering
\begin{tcolorbox}[colframe=gray, colback=lightgray!20, coltitle=black, sharp corners=southwest, rounded corners, width=0.9\textwidth, boxrule=0.5mm]
\textsf{\scriptsize  
Given a set of nodes with their coordinates, you need to find the shortest route that visits each node once and returns to the starting node. The task can be solved step-by-step by starting from the current node and iteratively choosing the next node. Help me design a novel algorithm that is different from the algorithms in the literature to select the next node in each step.
}
\end{tcolorbox}
\caption{Task description prompt used for the TSP.}
\label{fig:TSP_taskdescription}
\end{figure}

\begin{figure}[h!]
\centering
\begin{tcolorbox}[colframe=gray, colback=lightgray!20, coltitle=black, sharp corners=southwest, rounded corners, width=0.9\textwidth, boxrule=0.5mm]
\textsf{\scriptsize  
Given a set of customers and a fleet of vehicles with limited capacity, the task is to design a novel algorithm to select the next node in each step, with the objective of minimizing the total cost.
}
\end{tcolorbox}
\caption{Task description prompt used for the CVRP.}
\label{fig:CVRP_taskdescription}
\end{figure}

\begin{figure}[h!]
\centering
\begin{tcolorbox}[colframe=gray, colback=lightgray!20, coltitle=black, sharp corners=southwest, rounded corners, width=0.9\textwidth, boxrule=0.5mm]
\textsf{\scriptsize  
Implement a function that returns the priority with which we want to add an item to each bin.
}
\end{tcolorbox}
\caption{Task description prompt used for the OBP.}
\label{fig:OBP_taskdescription}
\end{figure}

\begin{figure}[h!]
\centering
\begin{tcolorbox}[colframe=gray, colback=lightgray!20, coltitle=black, sharp corners=southwest, rounded corners, width=0.9\textwidth, boxrule=0.5mm]
\textsf{\scriptsize  
Implement a novel heuristic strategy function that guides the lander in selecting actions step-by-step to achieve a safe landing. At each step, an appropriate action could be chosen based on the lander's current state and previous state, with the objective of reaching the target location in as few steps as possible. A 'safe landing' is defined as a touchdown with low vertical speed, upright orientation, and both angular velocity and angle close to zero, and both legs in contact with the ground.
}
\end{tcolorbox}
\caption{Task description prompt used for the LLC task.}
\label{fig:Lunar_taskdescription}
\end{figure}

\subsection{Prompt templates for evolutionary operators}
\label{operator_prompts}

In LES methods, different evolutionary operators are implemented through distinct prompt templates that guide the LLM toward specific search behaviors. To ensure fairness and isolate the contribution of the proposed mechanism rather than prompt engineering, DyCA adopts exactly the same operator prompts as EoH-S for the evolution of the complementary pool, including the Complementary-aware Search (CS) operator and the Local Search (LS) operator~\cite{liu2025eohs}, shown in Figures~\ref{fig:CS_prompt} and~\ref{fig:LS_prompt}.

In the specialist pool, we additionally introduce a Summary operator to accelerate specialization by explicitly distilling behavioral patterns from historically successful and unsuccessful algorithms~\cite{hu2025partition}. The prompt templates for generating summaries and for applying summarized guidelines are shown in Figures~\ref{fig:getsummary_prompt} and~\ref{fig:summary_prompt}, respectively.
For consistency, the same set of evolutionary operators is also employed in the specialist pool of InstSpecHH.

The initialization prompt used to generate the initial pool is shown in Figure~\ref{fig:init_prompt}.
Across all prompt templates, black text denotes fixed instructions, red placeholders indicate task-specific components, and blue placeholders represent algorithmic content that evolves throughout the evolutionary process.

\begin{figure}[h!]
\centering
\begin{tcolorbox}[colframe=gray, colback=lightgray!20, coltitle=black, sharp corners=southwest, rounded corners, width=0.9\textwidth, boxrule=0.5mm]
\textsf{\scriptsize 
\textcolor{red}{\textbf{``[Task Description Placeholder]''}} \\
Based on your expertise, please create a novel and efficient algorithm to solve this problem.\\
So far, experts have proposed \textcolor{blue}{\textbf{``[Value Placeholder]''}} algorithms. Their high-level ideas are summarized below.
The No. 1 algorithm and the corresponding code are:\\
\textcolor{blue}{\textbf{``[Algorithm Description Placeholder]''}} \\
\textcolor{blue}{\textbf{``[Code Placeholder]''}} \\
The No. 2 algorithm and the corresponding code are:\\
\textcolor{blue}{\textbf{``[Algorithm Description Placeholder]''}} \\
\textcolor{blue}{\textbf{``[Code Placeholder]''}} \\
...\\
Your goal is to design a new algorithm whose core idea differs from *all existing ones* by at least 30\%.\\
Please design a new algorithm following the instructions below:\\
1. Describe your concept for the new algorithm and its main steps in as few words as possible while ensuring clarity, and enclose it **exactly** within the markers '$\ll$' and '$\gg$', like this: $\ll$Your idea here$\gg$.\\
2. Implement your proposed algorithm using the following Python function template:\\
\textcolor{red}{\textbf{``[Code Template Placeholder]''}}}
\end{tcolorbox}
\caption{Prompt template for Initialization.}
\label{fig:init_prompt}
\end{figure}

\begin{figure}[h!]
\centering
\begin{tcolorbox}[colframe=gray, colback=lightgray!20, coltitle=black, sharp corners=southwest, rounded corners, width=0.9\textwidth, boxrule=0.5mm]
\textsf{\scriptsize 
\textcolor{red}{\textbf{``[Task Description Placeholder]''}} \\
I have \textcolor{blue}{\textbf{``[Value Placeholder]''}} existing algorithms with their codes as follows:\\
The No. 1 algorithm and the corresponding code are:\\
\textcolor{blue}{\textbf{``[Algorithm Description Placeholder]''}} \\
\textcolor{blue}{\textbf{``[Code Placeholder]''}} \\
The No. 2 algorithm and the corresponding code are:\\
\textcolor{blue}{\textbf{``[Algorithm Description Placeholder]''}} \\
\textcolor{blue}{\textbf{``[Code Placeholder]''}} \\
These algorithms are effective for solving different instance distributions. Please help me create a new algorithm that is different from the given ones. \\
1. First, describe your new algorithm and main steps in one concise sentence. Enclose your sentence **exactly** within the markers '$\ll$' and '$\gg$' like this: $\ll$Your idea here$\gg$.\\
2. Next, implement the algorithm in the following Python function:\\
\textcolor{red}{\textbf{``[Code Template Placeholder]''}}}
\end{tcolorbox}
\caption{Prompt template for the CS operator.}
\label{fig:CS_prompt}
\end{figure}

\begin{figure}[h!]
\centering
\begin{tcolorbox}[colframe=gray, colback=lightgray!20, coltitle=black, sharp corners=southwest, rounded corners, width=0.9\textwidth, boxrule=0.5mm]
\textsf{\scriptsize 
\textcolor{red}{\textbf{``[Task Description Placeholder]''}} \\
I have one algorithm with its code as follows.\\ 
Algorithm description:\\
\textcolor{blue}{\textbf{``[Algorithm Description Placeholder]''}} \\
Code:\\
\textcolor{blue}{\textbf{``[Code Placeholder]''}} \\
Please assist me in creating an improved version of the algorithm provided.\\
1. First, describe your new algorithm and main steps in one sentence. Enclose your sentence **exactly** within the markers '$\ll$' and '$\gg$' like this: $\ll$Your idea here$\gg$.\\
2. Next, implement the following Python function:\\
\textcolor{red}{\textbf{``[Code Template Placeholder]''}}}
\end{tcolorbox}
\caption{Prompt template for the LS operator.}
\label{fig:LS_prompt}
\end{figure}

\begin{figure}[h!]
\centering
\begin{tcolorbox}[colframe=gray, colback=lightgray!20, coltitle=black, sharp corners=southwest, rounded corners, width=0.9\textwidth, boxrule=0.5mm]
\textsf{\scriptsize 
You are an expert Algorithm Analyst and Optimization Coach. You are assisting in an algorithm design process to solve the following task:\\
\textcolor{red}{\textbf{``[Task Description Placeholder]''}} \\
I will provide you with two sets of algorithms generated so far:\\
1. **High Performing Group**: Algorithms that achieved high scores and efficiency.\\
2. **Low Performing Group**: Algorithms that performed poorly (but are syntactically correct).\\
Your Goal is to analyze the differences between these two groups to understand what makes an algorithm successful for this specific task.\\
Please provide a technical analysis report following this structure:\\
1.  **Pattern Analysis**:\\
    * What common logic or strategies are present in the *High Performing Group*?\\
    * What specific inefficiencies or logical flaws appear in the *Low Performing Group*?\\
2.  **Actionable Tips (The most important part)**:\\
    * Synthesize your analysis into 1-2 concise, actionable guidelines for generating future code.\\
    * Focus on "Do this" (to mimic success) and "Avoid this" (to prevent failure).\\
Here is the data for your analysis:\\
=== HIGH PERFORMING GROUP (Study these patterns) ===\\
High Performing Algorithm 1:\\
\textcolor{blue}{\textbf{``[Algorithm Description Placeholder]''}} \\
\textcolor{blue}{\textbf{``[Code Placeholder]''}} \\
...\\
=== LOW PERFORMING GROUP (Avoid these mistakes) ===\\
Low Performing / Failed Algorithm 1:\\
\textcolor{blue}{\textbf{``[Algorithm Description Placeholder]''}} \\
\textcolor{blue}{\textbf{``[Code Placeholder]''}} \\
...\\}
\end{tcolorbox}
\caption{Prompt template for getting summary.}
\label{fig:getsummary_prompt}
\end{figure}

\begin{figure}[h!]
\centering
\begin{tcolorbox}[colframe=gray, colback=lightgray!20, coltitle=black, sharp corners=southwest, rounded corners, width=0.9\textwidth, boxrule=0.5mm]
\textsf{\scriptsize 
Your task is:
\textcolor{red}{\textbf{``[Task Description Placeholder]''}} \\
I need you to evolve a new algorithm based on a "Parent Algorithm" and a set of "Optimization Guidelines".\\
=== PART 1: The Parent Algorithm ===\\
This is the current version of the algorithm and code. It functions, but we need to improve its score/efficiency.\\
Concept: \textcolor{blue}{\textbf{``[Algorithm Description Placeholder]''}} \\
Implementation: \textcolor{blue}{\textbf{``[Code Placeholder]''}} \\
=== PART 2: Optimization Guidelines (Crucial) === \\
We have analyzed the history of successful and failed attempts for this problem. You MUST incorporate the following advice into your new design: \textcolor{blue}{\textbf{``[Specialist Pool Summary Placeholder]''}}\\
=== PART 3: Your Task === \\
Please create an improved algorithm that: \\
1. Inherits the good logic from the Parent Algorithm.\\
2. Strictly applies the strategies mentioned in the "Optimization Guidelines".\\
3. Fixes any potential issues warned about in the guidelines.\\
**Output Format:**\\
First, describe your new algorithm and its core idea in **one concise sentence**. Enclose your sentence **exactly** within the markers '$\ll$' and '$\gg$' like this: $\ll$Your idea here$\gg$.\\
Next, implement the improved version using this Python function template:\\
\textcolor{red}{\textbf{``[Code Template Placeholder]''}}}
\end{tcolorbox}
\caption{Prompt template for the Summary operator.}
\label{fig:summary_prompt}
\end{figure}

\section{Licenses for used assets}
\label{Licenses_for_used_assets}
Since Funsearch is not officially open-source, we conducted our experiments using the version of Funsearch provided on the LLM4AD platform.

\begin{table}[ht]
\centering
\caption{Asset Usage and License Information}
\begin{tabular}{llll}
\toprule
Type                  & Asset     & License     & Usage      \\ \midrule
\multirow{3}{*}{Code} & Funsearch~\cite{romera2024mathematical} & MIT License & Evaluation \\
                      & EoH~\cite{liu2024evolution}       & MIT License & Evaluation \\
                      & ReEvo~\cite{ye2024reevo}     & MIT License & Evaluation \\
                      & EoH-S~\cite{liu2025eohs}     & MIT License & Evaluation \\\midrule
Dataset               & LLM4AD~\cite{liu2024llm4ad}    & MIT License & Testing    \\ \bottomrule
\end{tabular}

\end{table}




\end{document}